\pdfoutput=1
\documentclass{article}
\usepackage{arxiv}

\usepackage[utf8]{inputenc} 
\usepackage[T1]{fontenc}    
\usepackage{hyperref}       
\usepackage{url}            
\usepackage{booktabs}       
\usepackage{amsfonts}       
\usepackage{nicefrac}       
\usepackage{microtype}      
\usepackage{lipsum}
\usepackage{graphicx}
\graphicspath{ {./images/} }

\usepackage{multirow}
\usepackage{array}
\usepackage{amssymb}
\usepackage{pifont}
\newcommand{\cmark}{\ding{51}}%
\newcommand{\xmark}{\ding{55}}%
\usepackage{amsmath}
\usepackage{booktabs}
\usepackage{caption}
\usepackage{subfigure}
\usepackage{footnote}
\makesavenoteenv{tabular}
\makesavenoteenv{table}
\usepackage{tablefootnote}
\usepackage{makecell}
\title{A Generalization of ViT/MLP-Mixer to Graphs}

\author{%
  Xiaoxin He$^{1}$ 
  \hspace{0.18cm}
  Bryan Hooi$^1$
  \hspace{0.18cm}
  Thomas Laurent$^2$ 
  \hspace{0.18cm}
  Adam Perold$^{3}$ 
  \hspace{0.18cm}
  Yann LeCun$^{4.5}$ 
  \hspace{0.18cm}
  Xavier Bresson$^1$ \vspace{0.15cm} \\
  \texttt{\{xiaoxin, bhooi, xaviercs\}@comp.nus.edu.sg, tlaurent@lmu.edu} \\ \texttt{ap@elementresearch.com, yann@cs.nyu.edu } \vspace{0.15cm} \\
  {\normalfont $^1$National University of Singapore \hspace{0.18cm} $^2$Loyola Marymount University \hspace{0.18cm} $^3$Element, Inc.}\\
  $^4$New York University \hspace{0.18cm} $^5$Meta AI \hspace{0.14cm}
}

\usepackage{xspace}

\usepackage{color}

\usepackage[numbers]{natbib}

\begin{document}
\maketitle

\begin{abstract}
Graph Neural Networks (GNNs) have shown great potential in the field of graph representation learning. Standard GNNs define a local message-passing mechanism which propagates information over the whole graph domain by stacking multiple layers. This paradigm suffers from two major limitations, over-squashing and poor long-range dependencies, that can be solved using global attention but significantly increases the computational cost to quadratic complexity. 
In this work, we propose an alternative approach to overcome these structural limitations by leveraging the ViT/MLP-Mixer architectures introduced in computer vision.
We introduce a new class of GNNs, called Graph ViT/MLP-Mixer, that holds three key properties. First, they capture long-range dependency and mitigate the issue of over-squashing as demonstrated on Long Range Graph Benchmark and TreeNeighbourMatch datasets. Second, they offer better speed and memory efficiency with a complexity linear to the number of nodes and edges, surpassing the related Graph Transformer and expressive GNN models. Third, they show high expressivity in terms of graph isomorphism as they can distinguish at least 3-WL non-isomorphic graphs. 
We test our architecture on 4 simulated datasets and 7 real-world benchmarks, and show highly competitive results on all of them. 
The source code is available for reproducibility at: \url{https://github.com/XiaoxinHe/Graph-ViT-MLPMixer}.
\end{abstract}

\section{Message-Passing GNNs and the Limitations}\label{sec: intro}

In this first section, we present the background of the project by introducing the standard Message-Passing (MP) GNNs and their two major limitations; low expressivity representation and poor long-range dependency. We also present the current techniques that address these issues, i.e. Weisfeiler-Leman GNNs, graph positional encoding and Graph Transformers, as well as their shortcomings.

{\bf Message-Passing GNNs (MP-GNNs).} GNNs have become the standard learning architectures for graphs based on their  flexibility to work with complex data domains s.a. recommendation \citep{monti2017geometric,van2018graph}, chemistry \citep{duvenaud2015convolutional,gilmer2017neural}, physics \citep{cranmer2019learning, bapst2020unveiling}, transportation \citep{derrowpinion2021traffic}, vision \citep{han2022vision}, natural language processing (NLP) \citep{wu2021graph}, knowledge graphs \citep{schlichtkrull2018modeling}, drug design \citep{stokes2020deep, gaudelet2020utilising} and medical domain \citep{li2020graph,li2021braingnn}. Most GNNs are designed to have two core components. First, a structural message-passing mechanism s.a. \citet{defferrard2016convolutional, kipf2017semi, hamilton2017inductive, monti2017geometric, bresson2017gatedgcn, gilmer2017neural, velivckovic2018graph} that computes node representations by aggregating the local 1-hop neighborhood information. Second, a stack of $L$ layers that aggregates $L$-hop neighborhood nodes to increase the expressivity of the network and transmit information between nodes that are $L$ hops apart.

{\bf Weisfeiler-Leman GNNs (WL-GNNs).} One of the major limitations of MP-GNNs is their inability to distinguish (simple) non-isomorphic graphs. This limited expressivity can be formally analyzed with the Weisfeiler-Leman graph isomorphism test \citep{weisfeiler1968reduction}, as first proposed in \citet{xu2018how, morris2019weisfeiler}. Later on, \citet{maron2018invariant} introduced a general class of $k$-order WL-GNNs that can be proved to universally represent any class of $k$-WL graphs \citep{maron2019provably, chen2019equivalence}. But to achieve such expressivity, this class of GNNs requires using $k$-tuples of nodes with memory and speed complexities of $O(N^k)$, with $N$ being the number of nodes and $k\geq 3$. Although the complexity can be reduced to $O(N^2)$ and $O(N^3)$ respectively \citep{maron2019provably,chen2019equivalence,azizian2020expressive}, it is still computationally costly compared to the linear complexity $O(E)$ of MP-GNNs with $E$ being the number of edges, which often reduces to $O(N)$ for real-world graphs that exhibit sparse structures.
In order to reduce memory and speed complexities of WL-GNNs while keeping high expressivity, several works have focused on designing graph networks from their sub-structures s.a. sub-graph isomorphism \citep{bouritsas2022improving}, sub-graph routing mechanism \citep{alsentzer2020subgraph}, cellular WL sub-graphs \citep{bodnar2021weisfeiler}, and k-hop egonet sub-graphs~\citep{xu2018how, zhang2021nested, chen2019equivalence, zhao2021stars, sun}.

{\bf Graph Positional Encoding (PE).} Another aspect of the limited expressivity of GNNs is their inability to recognize simple graph structures s.a. cycles or cliques, which are often present in molecules and social graphs \citep{chen2020can}. We can consider $k$-order WL-GNNs with value $k$ to be the length of cycle/clique, but with high complexity $O(N^k)$. An alternative approach is to add positional encoding to the graph nodes. It was proved in~\citet{murphy2019relational,Loukas2020What} that unique and equivariant PE increases the representation power of any MP-GNN while keeping the linear complexity. 
This theoretical result was applied with great empirical success with index PE~\citep{murphy2019relational},  Laplacian eigenvectors~\citep{dwivedi2020benchmarking,dwivedi2021generalization,kreuzer2021rethinking,lim2022sign} and k-step Random Walk~\citep{li2020distance,dwivedi2021graph}.
All these graph PEs lead to GNNs strictly more powerful than the 1-WL test, which seems to be enough expressivity in practice \citep{zopf20221}. However, none of the PE proposed for graphs can provide a global position of the nodes that is unique, equivariant and distance sensitive. This is due to the fact that a canonical positioning of nodes does not exist for arbitrary graphs, as there is no notion of up, down, left and right on graphs. For example, any embedding coordinate system like graph Laplacian eigenvectors \citep{belkin2003laplacian} can flip up-down directions, right-left directions, and would still be a valid PE. This introduces ambiguities for the GNNs that require to (learn to) be invariant with respect to the graph or PE symmetries. A well-known example is given by the eigenvectors: there exist $2^k$ number of possible sign flips for $k$ eigenvectors that require to be learned by the network.

{\bf Issue of long-range dependencies.}
Another major limitations of MP-GNNs is the well-known issue of long-range dependencies. Standard MP-GNNs require $L$ layers to propagate the information from one node to their $L$-hop neighborhood. This implies that the receptive field size for GNNs can grow exponentially, for example with $O(2^L)$ for binary tree graphs. This causes over-squashing; exponentially growing information  is compressed into a fixed-length vector 
by the aggregation mechanism~\citep{alon2020bottleneck, ToppingGC0B22}. 
It is worth noting that the poor long-range modeling ability of deep GNNs can be caused by the combined effect of multiple factors, such as over-squashing, vanishing gradients, poor isomorphism expressivity, etc. but, in this work, we focus our effort on alleviating over-squashing s.a.~\citet{deac2022expander,arnaiz2022diffwire}.
Over-squashing is well-known since recurrent neural networks \citep{hochreiter1997long}, which have led to the development of the (self- and cross-)attention mechanisms for the translation task \citep{bahdanau2014neural,vaswani2017attention} first, and then for more general NLP tasks \citep{devlin2018bert,brown2020language}. 
Transformer architectures are the most elaborated networks that leverage attention, and have gained great success in NLP and computer vision (CV).
Several works have generalized the transformer architecture for graphs, alleviating the issue of long-range dependencies and achieving competitive or superior performance against standard MP-GNNs. 
We highlight the most recent research works in the next paragraph.

\textbf{Graph Transformers.} \citet{dwivedi2021generalization} generalize Transformers to graphs, with graph Laplacian eigenvectors as node PE, and incorporating graph structure into the permutation-invariant attention function. SAN and LSPE~\citep{kreuzer2021rethinking,dwivedi2021graph} further improve with PE learned from Laplacian and random walk operators. GraphiT~\citep{mialon2021graphit} encodes relative PE derived from diffusion kernels into the attention mechanism. GraphTrans~\citep{wu2021GraphTrans} add Transformers on the top of standard GNN layers. SAT~\citep{chen2022structure_SAT} proposes a novel self-attention mechanism that incorporates structural information into the standard self-attention module by using a GNN to compute subgraph representations.
Graphormer~\citep{ying2021graphormer} introduce three structural encodings, with great success on large molecular benchmarks. GPS~\citep{rampavsek2022recipe} categorizes the different types of PE and puts forward a hybrid MPNN+Transformer architecture. We refer to \citet{min2022transformer} for an overview of graph-structured Transformers. Generally, most Graph Transformer architectures address the problems of over-squashing and limited long-range dependencies in GNNs but they also increase significantly the complexity from $O(E)$ to $O(N^2)$, resulting in a computational bottleneck. A detailed description of related literature can be found in Appendix~\ref{app sec: related work}.

\section{Generalizing ViT/MLP-Mixer to Graphs}\label{sec:contribution}
In the following, we explain the importance of generalizing the ViT/MLP-Mixer architectures from CV to graphs. 

\textbf{ViT and MLP-Mixer in computer vision.} Transformers have gained remarkable success in CV and NLP, most notably with architectures like ViT~\citep{dosovitskiy2020ViT} and BERT~\citep{devlin2018bert}. The success of transformers has been long attributed to the attention mechanism~\citep{vaswani2017attention}, which is able to model long-range dependency by making "everything connected to everything".
But recently, this prominent line of networks has been challenged by more cost efficient alternatives. 
A novel family of models based on the MLP-Mixer introduced by \citet{tolstikhin2021mlp} has emerged and gained recognition for its simplicity and its efficient implementation. Overall, MLP-Mixer replaces the attention module with multi-layer perceptrons (MLPs) which are also not affected by over-squashing and poor long-range interaction. The original architecture is simple~\citep{tolstikhin2021mlp}; it takes image patches (or tokens) as inputs, encodes them with a linear layer (equivalent to a convolutional layer over the image patches), and updates their representations with a series of feed-forward layers applied alternatively to image patches (or tokens) and features. 
These plain networks~\citep{tolstikhin2021mlp, touvron2021resmlp, liu2021gmlp, wang2022dynamixer} can perform competitively with state-of-the-art (SOTA) vision Transformers, which tends to indicate that attention is not the only important inductive bias, but other elements like the general architecture of Transformers with patch embedding, residual connection and layer normalization, and carefully-curated data augmentation techniques seem to play essential roles as well~\citep{yu2022metaformer}.

\textbf{The benefits of generalizing ViT/MLP-Mixer from grids to graphs.} 
Standard MP-GNNs have linear learning/inference complexities but low representation power and poor long-range dependency. Graph Transformers address these two problems but loose the computational efficiency with a quadratic complexity price. A generalization of ViT/MLP-Mixer to graphs overcomes the computational bottleneck of Graph Transformers and solves the issue of long-distance dependency, hence going beyond standard MP-GNNs.  However, achieving such a successful generalization is challenging given the irregular and variable nature of graphs. See Section \ref{sec: challenges} for a detailed presentation of theses challenges.

\textbf{Contribution.} 
Our contributions are listed as follows. 
\begin{itemize}
\item We identify the key challenges to generalize ViT/MLP-Mixer from images to graphs and design a new efficient class of GNNs, namely Graph ViT/MLP-Mixer, that simultaneously captures long-range dependency, keeps linear speed/memory complexity, and achieves high graph isomorphic expressivity.
    \item We show competitive results on multiple benchmarks from Benchmarking GNNs~\citep{dwivedi2020benchmarking} and the Open Graph Benchmark (OGB)~\citep{hu2020open}; specifically, with 0.073 MAE on ZINC and 0.7997 ROCAUC on MolHIV.
     \item We demonstrate the capacity of the proposed model to capture long-range dependencies with SOTA performance on Long Range Graph Benchmark (LRGB)~\citep{dwivedi2022long} while keeping low complexity,  to mitigate the over-squashing issue on the TreeNeighbourMatch dataset~\citep{alon2020bottleneck}, and to reach the 3-WL expressive power on the SR25 dataset~\citep{balcilar2021breaking}. 
    \item Our approach forms a bridge between CV, NLP and graphs under a unified architecture, that can potentially benefit cross-over domain collaborations to design better networks.
\end{itemize}

\section{Generalization Challenges}\label{sec: challenges}

\begin{table*}[t]
    \caption{Differences between ViT/MLP-Mixer components for images and graphs. }
    
    \centering    
    \begin{tabular}{lll}
    \toprule
           & \multicolumn{1}{c}{Images}   & \multicolumn{1}{c}{Graphs} \\
         \midrule
           & 
         \begin{minipage}{.3\textwidth}
         \includegraphics[width=0.9\linewidth]{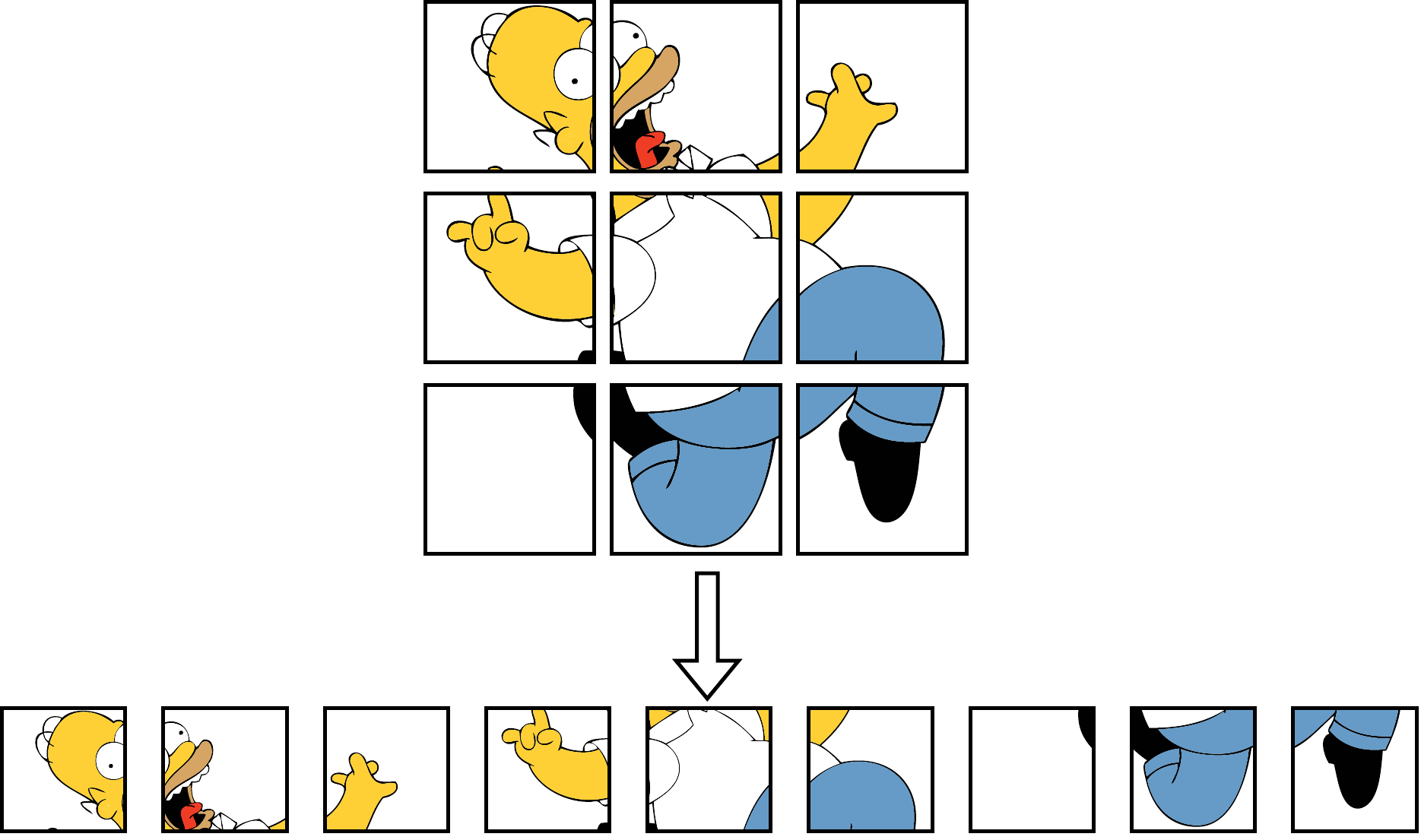}
         \end{minipage} & 
         \begin{minipage}{.3\textwidth}
         \centering
         \includegraphics[width=0.8\linewidth]{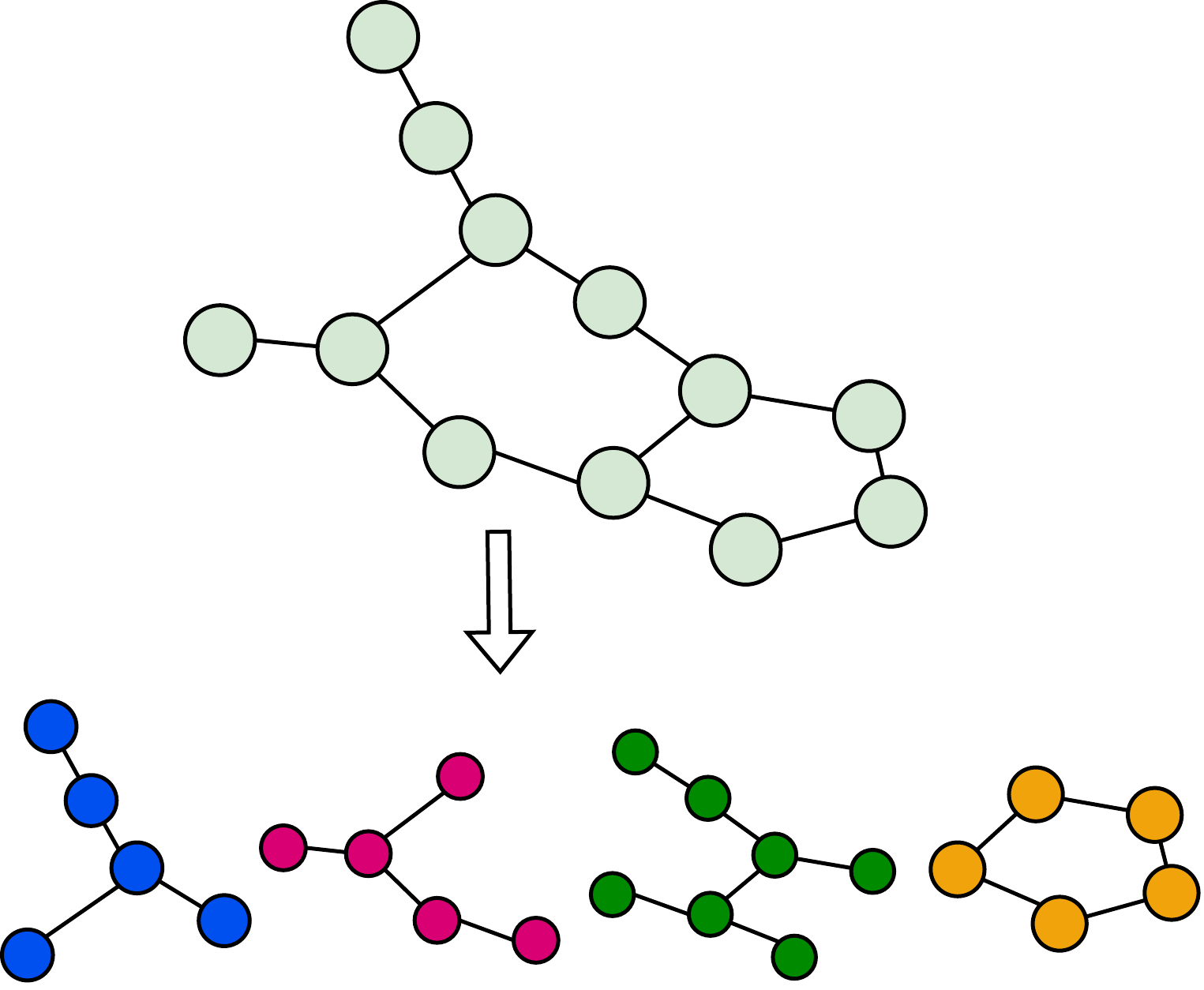}
         \end{minipage} 
         \\
         \midrule
         \multirow{3}{*}{Input} 
         & Regular grid & Irregular domain \\
         & Same data resolution &  Variable data structure  \\
         & \quad (Height, Width) & \quad (\# Nodes and \# Edges) \\
         \midrule
        \multirow{3}{*}{Patch Extraction}  
        & Via pixel reordering  & 
        {Via graph clustering algorithm} \\
        & Non-overlapping patches &  
        {Overlapping patches} \\
        & Same patches at each epoch  & 
        {Different patches at each epoch} \\
        
        \midrule
        \multirow{3}{*}{Patch Encoder}
        & Same patch resolution  & Variable patch structure \\
        & \quad (Patch Height, Patch Width)  & \quad (\# Nodes and \# Edges) \\
        & MLP (equivalently CNN) & GNN (e.g. GCN, GAT, GT) \\
        
        \midrule
        \multirow{3}{*}{Positional Information }
        & Implicitly ordered & No universal ordering\\
        & \quad (No need for explicit PE) & 
        {Node PE for patch encoder} \\
        & & 
        {Patch PE for token mixer} \\
        
        \midrule
        \multirow{2}{*}{ViT / MLP-Mixer} 
        & MLP / Channel mixer
        & MLP / Channel mixer\\
         & MHA / Token mixer  & 
         gMHA / Token mixer
        \\\bottomrule
    \end{tabular}
    \label{tab: comparision}
\end{table*}

We list the main questions when adapting MLP-Mixer from images to graphs in the following and in Table~\ref{tab: comparision}. 

\textbf{(1) How to define and extract graph patches/tokens?} One notable geometrical property that distinguishes graph-structured data from regular structured data, such as images and sequences, is that there does not exist in general a canonical grid to embed graphs. As shown in Table~\ref{tab: comparision}, images are supported by a regular lattice, which can be easily split into multiple grid-like patches (also referred to as tokens) of the same size via fast pixel reordering. However, graph data is irregular: the number of nodes and edges in different graphs is typically different. Hence, graphs cannot be uniformly divided into similar patches across all examples in the dataset. Finally, the extraction process for graph patches cannot be uniquely defined given the lack of canonical graph embedding. This raises the questions of how we identify meaningful graph tokens, and quickly extract them.

\textbf{(2) How to encode graph patches into a vectorial representation?} 
 Since images can be reshaped into patches of the same  size, they can be linearly encoded with an MLP, or equivalently with a convolutional layer with kernel size and stride values equal to the patch size. However, graph patches are not all the same size: they have variable topology with different number of nodes, edges and connectivity. Another important difference is the absence of a unique node ordering for graphs, which constrains the process to be invariant to node re-indexing for generalization purposes. In summary, we need a process that can transform graph patches into a fixed-length vectorial representation for arbitrary subgraph structures while being permutation invariant.

\textbf{(3) How to preserve positional information for nodes and graph patches?}  
As shown in Table~\ref{tab: comparision}, image patches in the sequence have implicit positions since image data is always ordered the same way due to its unique embedding in the Euclidean space. For instance, the image patch at the upper-left corner is always the first one in the sequence and the image patch at the bottom-right corner is the last one. On this basis, the token mixing operation of the MLP-Mixer is able to fuse the same patch information. However, graphs are naturally not-aligned and the set of graph patches are therefore unordered. We face a similar issue when we consider the positions of nodes within each graph patch. In images, the pixels in each patch are always ordered the same way; in contrast, nodes in graph tokens are naturally unordered. Thus, how do we preserve local and global positional consistency for nodes and graph patches?

\textbf{(4) How to reduce over-fitting for Graph ViT/MLP-Mixer?} 
ViT/MLP-Mixer architectures are known to be strong over-fitters~\citep{liu2021gmlp}. Most MLP-variants~\citep{tolstikhin2021mlp, touvron2021resmlp, wang2022dynamixer} first pre-train on large-scale datasets, and then fine-tune on downstream tasks, coupled with a rich set of data augmentation and regularization techniques, e.g. cropping, random horizontal flipping, RandAugment~\citep{cubuk2020randaugment}, mixup~\citep{zhang2017mixup}, etc. While data augmentation has drawn much attention in CV and NLP, graph data augmentation methods are not yet as effective, albeit interest and works on this topic~\citep{zhao2021data}. Variable number of nodes, edges and connectivity make graph augmentation challenging. Thus, how do we augment graph-structured data given this nature of graphs?

\section{Proposed Architecture}\label{sec: archi}

\subsection{Overview}\label{sec: overview}

\begin{figure}[t]
    \centering
    \includegraphics[width=1.0\textwidth]{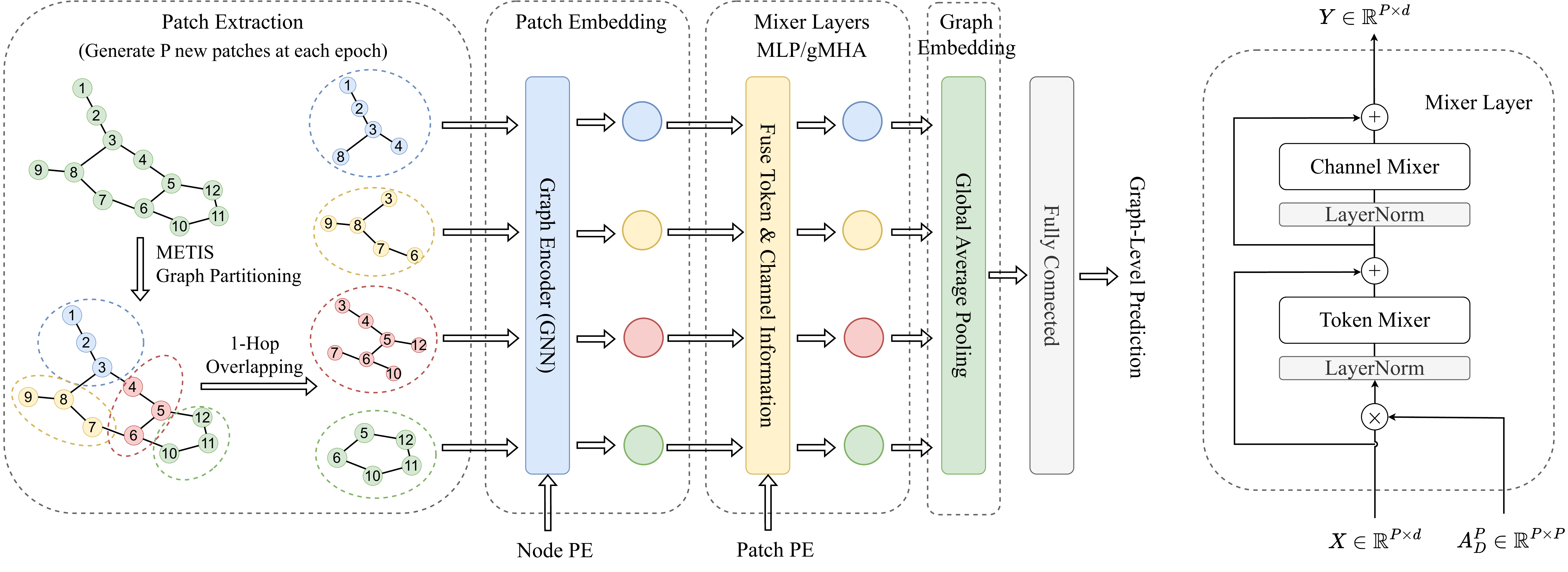}
    \caption{
    The basic architecture of the proposed Graph ViT/MLP-Mixer. They consist of a patch extraction module, a patch embedding module, a sequence of mixer layers, a global average pooling, and a classifier head. The patch extraction module partitions graphs into overlapping patches. The patch embedding module transforms these graph patches into corresponding token representations, which are fed into a sequence of mixer layers to generate the output tokens. A global average pooling layer followed by a fully-connected layer is finally used for prediction. Each Mixer Layer, MLP or graph-based multi-head attention (gMHA), is a residual network that alternates between a Token Mixer applied to all patches, and a Channel Mixer applied to each patch independently (see right side).
    }
    \label{fig: overview}
\end{figure}

The basic architecture of Graph MLP-Mixer is illustrated in Figure~\ref{fig: overview}. The goal of this section is to detail the choices we made to implement each component of the architecture. On the whole, these choices lead to a simple framework that provides speed and quality results.

\textbf{Notation.}  Let $G=(\mathcal{V}, \mathcal{E})$ be a graph with $\mathcal{V}$ being the set of nodes and $\mathcal{E}$ the set of edges. The graph has $N=|\mathcal{V}|$ nodes and $E = |\mathcal{E}|$ edges. The connectivity of the graph is represented by the adjacency matrix $A\in \mathbb{R}^{N \times N}$. The node features of node $i$ are denoted by $h_i$, while the features for an edge between nodes $i$ and $j$ are indicated by $e_{ij}$.  Let $\{\mathcal{V}_1, ..., \mathcal{V}_P\}$ be the nodes partition, $P$ be the pre-defined number of patches, and $G_i = (\mathcal{V}_i, \mathcal{E}_i)$ be the induced subgraph of $G$ with all the nodes in $\mathcal{V}_i$ and all the edges whose endpoints belong to $\mathcal{V}_i$. Let $h_G$ be the graph-level representation and $y_G$ be the graph-level target.

\subsection{Patch Extraction}
\label{sec:metis}

When generalizing MLP-Mixer to graphs, the first step is to extract patches. This extraction is straightforward for images. Indeed, all image data $x \in \mathbb{R}^{H\times W \times C}$ are defined on a regular grid with the same fixed resolution $(H,W)$, where $H$ and $W$ are respectively the height and the width, and $C$ is the number of channels. Hence, all images can be easily reshaped into a sequence of flattened patches $x_p\in \mathbb{R}^{P\times (R^2C)}$, where $(R,R)$ is the resolution of each image patch, and $P = HW/R^2$ is the resulting number of patches.

Unlike images with fixed resolution, extracting graph patches is more challenging, see Table~\ref{tab: comparision}. Generally, graphs have different sizes, i.e. number of nodes, and therefore cannot be uniformly divided like image data. Additionally, meaningful sub-graphs must be identified in the sense that nodes and edges composing a patch must share similar semantic or information, s.a. a community of friends sharing biking interest in a social network. As such, a graph patch extraction process must satisfy the following conditions: 1) the same extraction algorithm can be applied to any arbitrary graph, 2) the nodes in the sub-graph patch must be more closely connected than for those outside the patch, and 3) the extraction complexity must be fast, that is at most linear w.r.t. the number of edges, i.e. $O(E)$.

METIS~\citep{karypis1998metis} is a graph clustering algorithm with one of the best trade-off accuracy and speed. It partitions a graph into a pre-defined number of clusters such that the number of within-cluster links is much higher than between-cluster links in order to better capture good community structure. For these fine properties, we select it as our patch extraction algorithm. 
However, METIS is limited to finding non-overlapping clusters, as visualized in Figure~\ref{fig: overview}. In this example, METIS partitions the graph into four non-overlapping parts, i.e. $\{1,2,3\}, \{4,5,6\}, \{7,8,9\}$ and $\{10,11,12\}$, resulting in 5 edge cuts. Unlike images, extracting non-overlapping patches could imply losing important edge information, i.e. the cutting edges, and thus decreasing the predictive performance, as we will observe experimentally. To overcome this issue and to retain all original edges, we allow graph patches to overlap with each other. For example in Figure~\ref{fig: overview}, if the source and destination nodes of an edge are not in the same patch, we assign both nodes to the patches they belong to. As such, node 3 and node 4 are in two different patches, here the blue and red one, but are connected with each other. After our overlapping adjustment, these two nodes belong to both the blue and red patches. This practice is equivalent to expanding the graph patches to the one-hop neighbourhood of all nodes in that patch. Formally, METIS is first applied to partition a graph into $P$ non-overlapping patches: $\{\mathcal{V}_1, ..., \mathcal{V}_P\} \ \textrm{ such that } \mathcal{V} = \mathcal{V}_1 \cup ... \cup \mathcal{V}_P \  \textrm{ and } \  \mathcal{V}_i \cap \mathcal{V}_j = \emptyset, \  \forall i\ne j.$
Then, patches are expanded to their one-hop neighbourhood in order to preserve the information of between-patch links and make use of all graph edges: $\mathcal{V}_{i} \leftarrow  \mathcal{V}_{i} \ \cup \ \{ \ \mathcal{N}_1(j) \ | \ {j\in \mathcal{V}_i} \ \},$ where $\mathcal{N}_k(j)$ defines the $k$-hop neighbourhood of node $j$.

\subsection{Patch Encoder}

For images, patch encoding can be done with a simple linear transformation given the fixed resolution of all image patches. This operation is fast and well-defined. For graphs, the patch encoder network must be able to handle complex data structure such as invariance to index permutation, heterogeneous neighborhood, variable patch sizes, convolution on graphs, and expressive to differentiate graph isomorphisms. As a result, the graph patch encoder is a GNN, whose architecture is designed to best transform a graph token $G_p$ into a fixed-size representation $x_{G_p}$ into 3 steps.

\textbf{Step 1. Raw node and edge linear embedding.} The input node features $\alpha_i\in \mathbb{R}^{d_n \times 1} $ and edge features $\beta_{ij} \in \mathbb{R}^{d_e\times 1}$ are linearly projected into $d$-dimensional hidden features:
\begin{equation}
 h_i^0=U^0\alpha_i + u^0 \in\mathbb{R}^{d} ; \quad e_{ij}^0=V^0\beta_{ij}+v^0 \in\mathbb{R}^{d}
 \label{eq: input features}
\end{equation}
where $U^0 \in \mathbb{R}^{d\times d_n}$, $V^0\in\mathbb{R}^{d\times d_e}$ and $u^0, v^0\in \mathbb{R}^d$ are learnable parameters.

\textbf{Step 2. Graph convolutional layers with MP-GNN.} We apply a series of $L$ convolution layers, where the node and edge representations are updated with a MP-GNN applied to each graph patch $G_p=(\mathcal{V}_p, \mathcal{E}_p)$ as follows:
\begin{equation}
    \begin{split}
        &h_{i,p}^{\ell+1} = f_\textrm{node}(h_{i,p}^\ell, \{h_{j,p}^\ell|{j\in\mathcal{N}(i)}\}, e_{ij,p}^\ell) + g_\textrm{patch-node}(h_p^\ell),
         \quad h_{i,p}^{\ell+1}, h_{i,p}^{\ell},h_p^\ell \in \mathbb{R}^d,\\
    &e_{ij, p}^{\ell+1}=f_\textrm{edge}(h_{i,p}^\ell, h_{i,p}^\ell, e_{ij, p}^{\ell}) + g_\textrm{patch-edge}(e_p^\ell),
    \quad e_{ij, p}^{\ell+1}, e_{ij, p}^{\ell},e_p^\ell \in \mathbb{R}^d,
    \end{split}
\end{equation}
where $\ell$ is the layer index, $p$ is the patch index, $i,j$ denotes the nodes, $\mathcal{N}(i)$ is the neighborhood of the node $i$ and functions $f_\textrm{node}$ and $f_\textrm{edge}$ (with learnable parameters) define any arbitrary MP-GNN architecture s.a. \citep{kipf2017semi,bresson2017gatedgcn,hu2019gine,dwivedi2021generalization}, $h_p^\ell=\frac{1}{|\mathcal{V}_p|}\sum_{i\in \mathcal{V}_p} h_{i, p}^{l}\in\mathbb{R}^{d}$, $e_p^\ell=\frac{1}{|\mathcal{E}_p|}\sum_{ij\in \mathcal{E}_p} e_{ij, p}^{l}\in\mathbb{R}^{d}$ are respectively the mean representations of the patch nodes and patch edges, and $g_\textrm{patch-node}$, $g_\textrm{patch-edge}$ are MLP-based functions that act on $h_p^\ell$ and $e_p^\ell$. 
For each node and edge covered by more than one  patch due to the patch overlapping to include all edges cut by METIS, we update the node/edge representation by averaging over the overlapping patches:
\begin{eqnarray}
h_{i,p}^{l+1} &\leftarrow & \mathop{\textrm{Mean}}\limits_{\{k | i\in \mathcal{V}_k\}} h_{i,k}^{l+1} \in\mathbb{R}^{d},\\
e_{ij,p}^{l+1} &\leftarrow & \mathop{\textrm{Mean}}\limits_{\{k | ij\in \mathcal{E}_k\}} e_{ij,k}^{l+1} \in\mathbb{R}^{d},
\end{eqnarray}
where $\{k | i\in \mathcal{V}_k\}$, $\{k | ij\in \mathcal{E}_k\}$ are the set of all patches that cover node $i$, edge $ij$ respectively.

\textbf{Step 3. Pooling and readout.} The final step is mean pooling all node vectors in $G_p$ such that $h_p=\frac{1}{|\mathcal{V}_p|}\sum_{i\in \mathcal{V}_p} h_{i, p}^{\ell=L}\in\mathbb{R}^{d}$, and applying a small MLP to get the fixed-sized patch embedding $x_{G_p}\in\mathbb{R}^{d}$.

Observe that the patch encoder is a MP-GNN, and thus has the inherent limitation of poor long-range dependency. Does it affect the Graph MLP-Mixer to capture long-range interactions? The answer is negative because this problem is limited to large graphs. But for small patch graphs, this issue does not really exist (or is negligible). To give a few numbers, the mean number of nodes and the mean diameter for graph patches are around 3.15 and 1.82 respectively for molecular datasets and around 17.20 and 3.07 for image datasets, see Table \ref{tab: graph patch statistics}.

\subsection{Positional Information}
Regular grids offer a natural implicit arrangement for the sequence of image patches and for the pixels inside the image patches. However, such ordering of nodes and patches do not exist for general graphs. This lack of positional information reduces the expressivity of the network. Hence, we use two explicit positional encodings (PE); one absolute PE for the patch nodes and one relative PE for the graph patches.

\textbf{Node PE.} Input node features in Eq~(\ref{eq: input features}) are augmented with $p_i\in \mathbb{R}^K$, with a learnable matrix $T^0 \in \mathbb{R}^{d\times K}$:
\begin{equation}
 h_i^0=T^0 p_i + U^0\alpha_i + u^0 \in\mathbb{R}^{d},
 \label{eq: nodePE}
\end{equation}The benefits of different PEs are dataset dependent. We follow the strategy in~\citep{rampavsek2022recipe} that uses random-walk structural encoding (RWSE) \citep{dwivedi2021graph} for molecular data and Laplacian eigenvectors encodings \citep{dwivedi2020benchmarking} for image superpixels. Since Laplacian eigenvectors are defined up to  sign flips, the sign of the eigenvectors is randomly flipped during training.

\textbf{Patch PE.} Relative positional information between the graph patches can be computed from the original graph adjacency matrix $A\in \mathbb{R}^{N \times N}$ and the clusters $\{\mathcal{V}_1, ..., \mathcal{V}_P\}$ extracted by METIS in Section \ref{sec:metis}. Specifically, we capture relative positional information via the coarsened adjacency matrix $A^{P} \in \mathbb{R}^{P\times P}$ over the patch graphs: 
\begin{equation}
    A^{P}_{ij}=|\mathcal{V}_i \cap \mathcal{V}_j | =\textrm{Cut}(\mathcal{V}_i, \mathcal{V}_j),
    \label{eq: coarsen_adj}
\end{equation}

where $\textrm{Cut}(\mathcal{V}_i, \mathcal{V}_j)=\sum_{k\in \mathcal{V}_i} \sum_{l\in \mathcal{V}_j} A_{kl}$ is the standard graph cut operator which counts the number of connecting edges between cluster $\mathcal{V}_i$ and cluster $\mathcal{V}_j$. 

We extract the positional encoding $\hat p_i\in\mathbb{R}^{\hat K}$ at the patch level, similar to the node level, which will be injected (after a linear transformation) into the first layer of the mixer block:
\begin{equation}
    x_i^0 =  \hat T^0 \hat p_i  + \hat U^0 x_i + \hat u^0 \in \mathbb{R}^d,
    \label{eq: patch pe}
\end{equation}
where $x_i$ is the patch embedding.

\subsection{Mixer Layer}\label{subsec: mixer layer}
For images, the original mixer layer~\citep{tolstikhin2021mlp} is a simple network that alternates channel and token mixing steps. The token mixing step is performed over the token dimension, while the channel mixing step is carried out over the channel dimension. These two interleaved steps enable information fusion among tokens and channels. 
The simplicity of the mixer layer has led to a significant reduction in computational cost with little or no sacrifice in performance. Indeed, the self-attention mechanism in ViT requires $O(P^2)$ memory and $O(P^2)$ computation, while the mixer layer in MLP-Mixer needs $O(P)$ memory and $O(P)$ computation. 

Let $X\in \mathbb{R}^{P\times d}$ be the patch embedding $\{x_{G_1},..., x_{G_P}\}$, the graph mixer layer can be expressed as 
\begin{equation}
    \begin{split}
        &U  = X + (W_2\sigma(W_1\ \textrm{LayerNorm}(X))) \in \mathbb{R}^{P\times d}
         \quad\quad\quad \textrm { Token mixer},\\
    &Y = U + (W_4\sigma(W_3 \textrm{LayerNorm}(U)^T))^T \in \mathbb{R}^{P\times d}
     \quad\quad\quad\ \textrm { Channel mixer},
    \end{split}
\end{equation}
where $\sigma$ is a GELU nonlinearity~\citep{hendrycks2016gaussian}, $\textrm{LayerNorm}(\cdot)$ is layer normalization  \citep{ba2016layer}, and matrices $W_1 \in \mathbb{R}^{d_s \times P}, W_2\in \mathbb{R}^{P \times d_s}$, $W_3\in \mathbb{R}^{d_c \times d}, W_4\in \mathbb{R}^{d\times d_c}$, where $d_s$ and $d_c$ are the tunable hidden widths in the token-mixing and channel-mixing MLPs.

Alternatively, we can formulate a graph transformer layer to incorporate the self-attention mechanism as in ViT:
\begin{equation}
\label{eq:gViT}
    \begin{split}
        &U  = X + \textrm{gMHA}(\textrm{LayerNorm}(X)) \in \mathbb{R}^{P\times d}, \\
    &Y = U + \textrm{MLP}(\textrm{LayerNorm}(U)) \in \mathbb{R}^{P\times d},
    \end{split}
\end{equation}
where $\textrm{gMHA}$ (graph-based multi-head attention) is designed to capture token dependencies based on the given graph topology. In Eq.\eqref{eq:gViT}, gHMA is defined as $\big(A^P\odot\textrm{softmax}(\frac{QK^T}{\sqrt{d}})\big)V$ but other options are possible to characterize the gHMA mechanism, as studied in Appendix~\ref{section: design of MHA}.

Then we generate the final graph-level representation by mean pooling all the non-empty patches:
\begin{equation}
    h_G = \sum_{p} m_p \cdot x_{G_p} / \sum_{p} m_p \  \in\mathbb{R}^{d},
\end{equation}
where $m_p$ is a binary variable with value 1 for non-empty patches and value 0 for empty patches (since graphs have variable sizes, small graphs can produce empty patches). 
Finally, we apply a small MLP to get the graph-level target:
\begin{equation}
y_{G} = \textrm{MLP}(h_G) \in \mathbb{R} \textrm{ (regression) } \quad \textrm{or} \quad \mathbb{R}^{n_c} \textrm{ (classification)}. 
\end{equation}

\subsection{Data augmentation}\label{subsec: data augmentation}
MLP-Mixer architectures are known to be strong over-fitters~\citep{liu2021gmlp}. 
In order to reduce this effect, we propose to introduce some perturbations in METIS as follows. Let $G=(\mathcal{V}, \mathcal{E})$ be the original graph and $G'=(\mathcal{V}, \mathcal{E}')$ be the graph after randomly dropping a small set of edges. At each epoch, we apply METIS graph partition algorithm on $G'$ to get slightly different node partitions $\{\mathcal{V}_1, ..., \mathcal{V}_P\}$. Then, we extract the graph patches $\{G_1, ..., G_P\}$ where $G_i = (\mathcal{V}_i, \mathcal{E}_i)$ is the induced subgraph of the original graph $G$, and not the modified $G'$. This way, we can produce distinct graph patches at each epoch that retain all the nodes and edges from the original graph.

\section{Experiments}
{\bf Graph Benchmark Datasets.} 
We evaluate our Graph ViT/MLP-Mixer on a wide range of graph benchmarks; 1) \textbf{Simulated datasets:} CSL, EXP, SR25 and TreeNeighbourMatch dataset,
2) \textbf{Small real-world datasets:} ZINC, MNIST and CIFAR10 from Benchmarking GNNs~\citep{dwivedi2020benchmarking}, and MolTOX21 and MolHIV from OGB~\citep{hu2020open} and 3) \textbf{Large real-world datasets:} Peptides-func and Peptides-struct from LRGB~\citep{dwivedi2022long}. Details are provided in Appendix~\ref{app_sec: Datasets description} and Appendix~\ref{app: experiment}.

\subsection{Comparison with MP-GNNs}

\begin{table*}[t]
\scriptsize
    \centering
    \caption{Comparison with base MP-GNNs. Results are averaged over 4 runs with 4 different seeds.}
     \label{tab: performance}
    \begin{tabular}{lccccccc}
    \toprule
         \multirow{2}{*}{Model} 
         &  ZINC  
         & MNIST  
         & CIFAR10  
         & MolTOX21  
         & MolHIV  
         & {Peptide-func} 
         & {Peptide-struct} 
         \\ 
         \cmidrule(lr){2-2}
         \cmidrule(lr){3-3}
         \cmidrule(lr){4-4}
         \cmidrule(lr){5-5}
         \cmidrule(lr){6-6}
         \cmidrule(lr){7-7}
         \cmidrule(lr){8-8}
         & MAE $\downarrow$ 
         & Accuracy $\uparrow$ 
         & Accuracy $\uparrow$
         & ROCAUC $\uparrow$ 
         & ROCAUC $\uparrow$ 
         & {AP $\uparrow$}  
         & {MAE $\downarrow$}\\
    \midrule
    GCN 
    & 0.1952 ± 0.0057 
    & 0.9269 ± 0.0023 
    & 0.5423 ± 0.0056 
    & 0.7525 ± 0.0031 
    & 0.7813 ± 0.0081 
    & 0.6328 ± 0.0086 
    & 0.2758 ± 0.0012 
    \\

    {GCN-MLP-Mixer} 
    & {\bf {0.1347 ± 0.0020}} 
    & {{0.9516 ± 0.0027}}          
    & {{0.6111 ± 0.0017}}           
    & {{0.7816 ± 0.0075}}          
    & {\bf{0.7929 ± 0.0111}}           
    & {{0.6832 ± 0.0061}}          
    & {{0.2486 ± 0.0041}}           
    \\

    {GCN-ViT} 
    & {0.1688 ± 0.0095}    
    & \bf{0.9600 ± 0.0015}    
    & \bf{0.6367 ± 0.0027}    
    & \bf{0.7820 ± 0.0096}    
    & {0.7780 ± 0.0120}    
    & \bf{0.6855 ± 0.0049}    
    & \bf{0.2468 ± 0.0015}    
    \\
    \midrule
    GatedGCN 
    & 0.1577 ± 0.0046 
    & 0.9776 ± 0.0017 
    & 0.6628 ± 0.0017 
    & 0.7641 ± 0.0057 
    & 0.7874 ± 0.0119 
    & 0.6300 ± 0.0029 
    & 0.2778 ± 0.0017 
    \\

    {GatedGCN-MLP-Mixer}
    & {{0.1244 ± 0.0053}} 
    & {{0.9832 ± 0.0004}}          
    & {{0.7060 ± 0.0022}}           
    & {\bf{0.7910 ± 0.0040}}          
    & {\bf{0.7976 ± 0.0136}}           
    & {{0.6932 ± 0.0017}}          
    & {{0.2508 ± 0.0007}}           
    \\
    
    {GatedGCN-ViT} 
    & \bf{0.1421 ± 0.0031}    
    & \bf{0.9846 ± 0.0009}    
    & \bf{0.7158 ± 0.0009}    
    & {0.7857 ± 0.0028}    
    & {0.7734 ± 0.0114}    
    & \bf{0.6942 ± 0.0075}    
    & \bf{0.2465 ± 0.0015}    
    \\
    \midrule
    
    GINE 
    & 0.1072 ± 0.0037 
    & 0.9705 ± 0.0023 
    & 0.6131 ± 0.0035 
    & 0.7730 ± 0.0064 
    & 0.7885 ± 0.0034 
    & 0.6405 ± 0.0077 
    & 0.2780 ± 0.0021 
    \\

    {GINE-MLP-Mixer}
    & {\bf {0.0733 ± 0.0014}} 
    & {{0.9809 ± 0.0004}}          
    & {{0.6833 ± 0.0022}}           
    & {\bf{0.7868 ± 0.0043}}          
    & {\bf{0.7997 ± 0.0102}}           
    & {\bf{0.6970 ± 0.0080}}          
    & {{0.2494 ± 0.0007}}           
    \\

    {GINE-ViT} 
    & {0.0849 ± 0.0047}    
    & \bf{0.9820 ± 0.0005}    
    & \bf{0.6967 ± 0.0040}    
    & {0.7851 ± 0.0077}    
    & {0.7792 ± 0.0149}    
    & {0.6919 ± 0.0085}    
    & \bf{0.2449 ± 0.0016}    
    \\
    \midrule
    
    GraphTrans
    & 0.1230 ± 0.0018
    & 0.9782 ± 0.0012
    & 0.6809 ± 0.0020
    & 0.7646 ± 0.0055
    & 0.7884 ± 0.0104
    & 0.6313 ± 0.0039
    & 0.2777 ± 0.0025
    \\

    GraphTrans-MLP-Mixer
    & {\bf {0.0773 ± 0.0030}} 
    & {\bf {0.9742 ± 0.0011}}          
    & {\bf {0.7396 ± 0.0033}}           
    & {{0.7817 ± 0.0040}}          
    & {\bf{0.7969 ± 0.0061}}           
    & {{0.6858 ± 0.0062}}          
    & {{0.2480 ± 0.0013}}           
    \\
    
    {GraphTrans-ViT} 
    & {0.0960 ± 0.0073}    
    & {0.9725 ± 0.0023}    
    & {0.7211 ± 0.0055}    
    & \bf{0.7835 ± 0.0032}    
    & {0.7755 ± 0.0208}    
    & \bf{0.6876 ± 0.0059}    
    & \bf{0.2455 ± 0.0027}    
    \\
    \bottomrule
    \end{tabular}
\end{table*}

We show in Table~\ref{tab: performance} that ViT/Graph MLP-Mixer lifts the performance of all base MP-GNNs across various datasets, which include GCN~\citep{kipf2017semi}, GatedGCN~\citep{bresson2017gatedgcn}, GINE~\citep{hu2019gine} and Graph Transformer~\citep{dwivedi2021generalization}. We augmented all the base models with the same type of PE as Graph MLP-Mixer to ensure a fair comparison. These promising results demonstrate the generic nature of our proposed architecture which can be applied to any MP-GNNs in practice. 
Remarkably, Graph ViT/MLP-Mixer outperforms the base MP-GNNs by large margins on two LRGB~\citep{dwivedi2022long} datasets; we can observe an average 0.056 Average Precision improvement on Peptides-func and an average 0.028 MAE decrease on Peptides-struct, which verifies its superiority over MP-GNNs in capturing long-range interaction.

\subsection{Comparison with SOTAs}

Next, we compare Graph ViT/MLP-Mixer against popular GNN models with SOTA results, including Graph Transformers (GraphiT, GPS, SAN, etc.) and expressive GNNs (GNN-AK+ and SUN),
as shown in Table~\ref{tab: sota} and Table~\ref{app tab: lrgb}. For small molecular graphs, our model achieved 0.073 on ZINC and 0.7997 on MolHIV. For larger molecular graphs, our model sets new SOTA performance with the best scores of 0.6970 on Peptides-func and 0.2449 on Peptides-struct.

Besides, Graph ViT/MLP-Mixer offers better space-time complexity and scalability. 
Theoretically, most Graph Transformer models and expressive GNNs, might be computationally infeasible for large graphs, as they need to calculate the full attention and need to run an inner GNN on every node of the graph respectively. Experimentally, we observed that, when training on datasets with hundreds of nodes, SAN+LapPE~\citep{chen2022structure_SAT} and SUN~\citep{sun} require $9.4\times$ and $43.8\times$ training time per epoch, and $12.4\times$ and $18.8\times$ memory respectively, compared to our model. 

\begin{table}[t]
    \centering
    \caption{Comparison of our best results from Table~\ref{tab: performance} with the state-of-the-art models (missing values from literature are indicated with '-').  Results are averaged over 4 runs with 4 different seeds. 
    }
    \label{tab: sota}
    \scriptsize
    \begin{tabular}{lcccccccc}
    \toprule
         \multirow{2}{*}{Model} 
         & ZINC 
         & MolHIV 
         & \multicolumn{3}{c}{Peptides-func} 
         & \multicolumn{3}{c}{Peptides-strcut} \\
         \cmidrule(lr){2-2}
         \cmidrule(lr){3-3}
         \cmidrule(lr){4-6}
         \cmidrule(lr){7-9}
           & MAE $\downarrow$  
           & ROCAUC $\uparrow$ 
           & {AP $\uparrow$ } 
           & {Time} 
           & {Memory } 
           & {MAE $\downarrow$ }       
           & {Time} 
           & {Memory} 
           \\
         \midrule
         GT~\citep{dwivedi2020benchmarking} & 
         0.226 ± 0.014 
         & -- 
         & --& --& --
         & --& --& --\\
         GraphiT~\citep{mialon2021graphit} & 
         0.202 ± 0.011 
         & -- 
         & --& --& --
         & --& --& --\\
         Graphormer~\citep{ying2021graphormer} 
         & 0.122 ± 0.006 
         & -- 
         & --& --& --
         & --& --& --\\
         GPS~\citep{rampavsek2022recipe}
         & \bf 0.070 ± 0.004 
         & 0.7880 ± 0.0101 
         & 0.6562 ± 0.0115 
         & $1.4\times$
         & $6.8\times$
         & 0.2515 ± 0.0012  
         & $1.3\times$
         & $8.3\times$\\
         SAN+LapPE~\citep{kreuzer2021rethinking} 
         & 0.139 ± 0.006 
         & 0.7775 ± 0.0061 
         & 0.6384 ± 0.0121 
         & $9.4\times$ & $12.4\times$
         & 0.2683 ± 0.0043 
         & $8.8\times$ &  $14.7\times$ \\
         SAN+RWSE~\citep{kreuzer2021rethinking} 
         & --
         & --
         & 0.6439 ± 0.0075 
         & $8.0\times$ & $19.5\times$ 
         & 0.2545 ± 0.0012 
         & $7.9\times$ &  $14.5\times$\\
         \midrule
         GNN-AK+~\citep{zhao2021stars}
         & {0.080 ± 0.001} 
         & {0.7961 ± 0.0119} 
         &  0.6480 ± 0.0089 
         & $2.6\times$& $7.8\times$
         & 0.2736 ± 0.0007  
         & $2.5\times$& $9.2\times$ \\
         
         SUN~\citep{sun}
         & {0.084 ± 0.002} 
         & {0.8003 ± 0.0055}\tablefootnote{For SUN,  we run the official code and obtain 0.7886 ± 0.0081 on MolHIV with our 4 seeds.} 
         & 0.6730 ± 0.0078  
         & $43.8\times$ & $18.8\times$
         &  0.2498 ± 0.0008 
         & $42.7\times$ & $20.7\times$\\
         
         CIN~\citep{bodnar2021weisfeiler}
         & {0.079 ± 0.006}~\tablefootnote{For CIN, the reporting score is not obtained with the budget of 500k parameters but with 1.7M parameters (3x more) when running their official code.} 
         & {0.8094 ± 0.0057}
         & {--}& {--}& {--}
         & {--}& {--}& {--}\\

         \midrule
         Graph MLP-Mixer
         & {0.073 ± 0.001} 
         & \bf {0.7997 ± 0.0102} 
         & \bf{0.6970 ± 0.0080} & $1.0\times$ &  $1.0\times$ 
         & 0.2475 ± 0.0015 & $1.0\times$ &  $1.2\times$ \\
         Graph ViT
         & {0.085 ± 0.005} 
         & {0.7792 ± 0.0149} 
         & {0.6942 ± 0.0075} & $1.1\times$  & $0.8\times$ 
         & \bf{0.2449 ± 0.0016} & $1.0\times$ &  $1.0\times$ 
         \\
         \bottomrule
    \end{tabular}  
\end{table}

\subsection{Graph ViT/MLP-Mixer can mitigate over-squashing}\label{sec: oversquashing}

TreeNeighbourMatch is a synthetic dataset proposed by~\citet{alon2020bottleneck} to provide an intuition into over-squashing. Each example is a binary tree of depth $r$. The goal is to predict an alphabetical label for the target node, where the correct answer is the label of the leaf node that has the same degree as the target node. Figure~\ref{fig: tree neighbour} shows that standard MP-GNNs (i.e., GCN, GGCN, GAT and GIN) fail to generalize on the dataset from $r=4$, whereas
our model mitigates over-squashing and generalizes well until $r=7$.
As for why this happens, \citet{alon2020bottleneck} show that GNNs fail to solve larger TreeNeighbourMatch cases as they `squash' information about the graph into the target node's embedding, which can hold a limited amount of information. In contrast, Graph ViT/MLP-Mixer avoids this problem as it transmits long-range information directly without `squashing.' Concretely, Appendix \ref{app_sec: oversquashing} shows a simple construction illustrating that our model can solve TreeNeighbourMatch cases while avoiding the inherent limitations of MP-GNNs discussed by \citet{alon2020bottleneck}.

  \begin{minipage}{\textwidth}
  \begin{minipage}[c]{0.48\textwidth}
    \centering
      \includegraphics[scale=0.25]{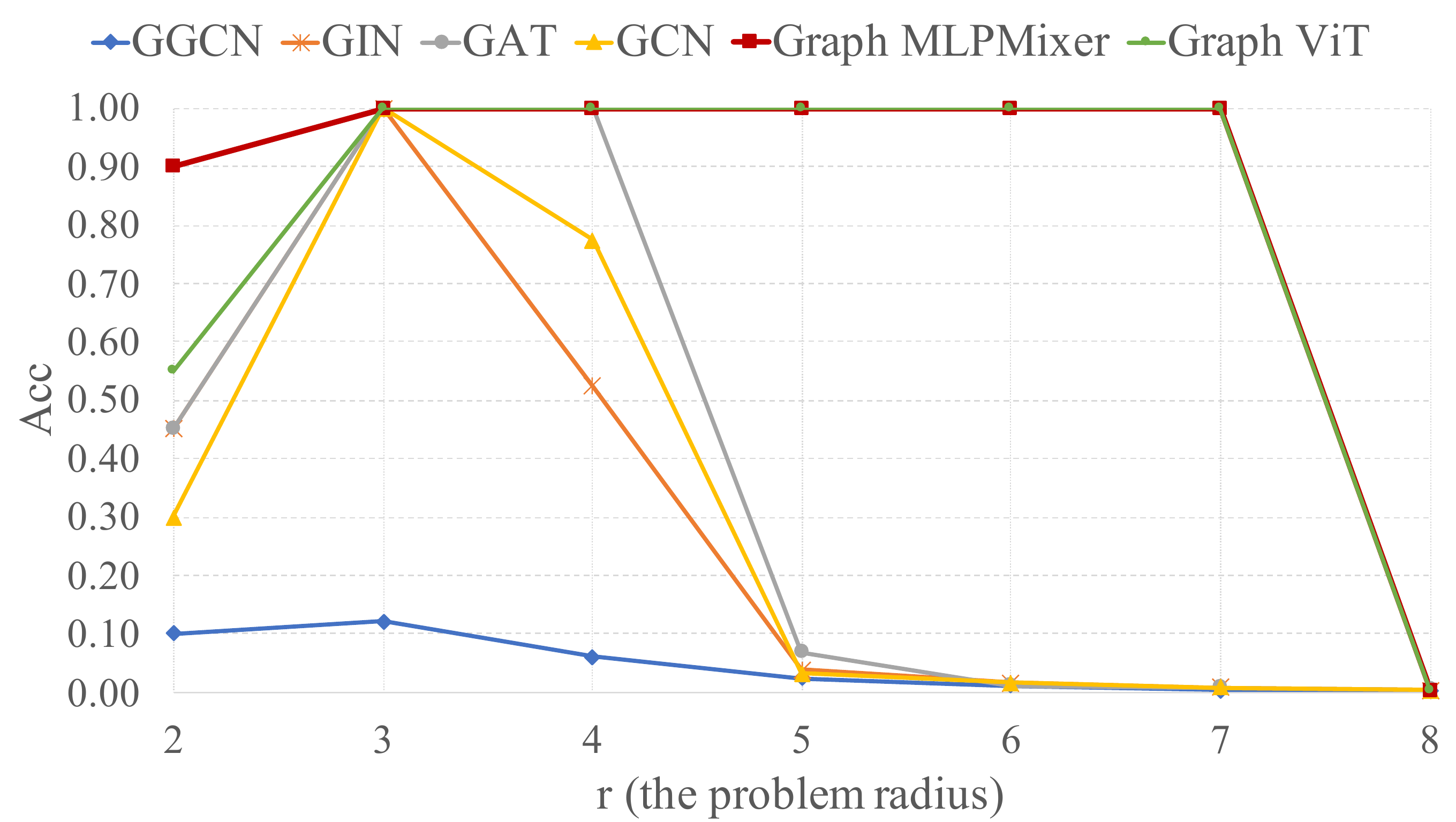}
      \captionsetup{type=figure}
    \caption{Test Accuracy across problem radius (tree depth) in the TreeNeighbourMatch problem.}
    \label{fig: tree neighbour}
  \end{minipage}
  \hfill
  \begin{minipage}[c]{0.50\textwidth}
  \captionsetup{type=table}
  \caption{Empirical evaluation of the expressive power on three simulation datasets, averaging over 4 runs with 4 different seeds.}
  \label{tab: Expressivity}
  \scriptsize
    \centering
    \begin{tabular}{lccc}
    \toprule
    Model & CSL (ACC)& EXP (ACC) & SR25 (ACC) \\
    \midrule
    GCN & 10.00 ± 0.00 & 51.90 ± 1.96 & 6.67 ± 0.00\\
    GatedGCN & 10.00 ± 0.00 & 51.73 ± 1.65& 6.67 ± 0.00\\
    GINE & 10.00 ± 0.00 & 50.69 ± 1.39&6.67 ± 0.00\\
    GraphTrans& 10.00 ± 0.00 & 52.35 ± 2.32 &6.67 ± 0.00\\
    \midrule
         {GCN-MLP-Mixer} & 100.00 ± 0.00 & 100.00 ± 0.00 & 100.00 ± 0.00  \\
         {GatedGCN-MLP-Mixer} & 100.00 ± 0.00 & 100.00 ± 0.00 & 100.00 ± 0.00  \\
         {GINE-MLP-Mixer} & 100.00 ± 0.00 & 100.00 ± 0.00 & 100.00 ± 0.00  \\
         {GraphTrans-MLP-Mixer}& 100.00 ± 0.00 & 100.00 ± 0.00 & 100.00 ± 0.00  \\
         \bottomrule
    \end{tabular}
    \end{minipage}
  \end{minipage}
\subsection{Graph ViT/MLP-Mixer can achieve empirical high expressivity}

%

We experimentally evaluate the expressive power of Graph ViT/MLP-Mixer on three simulated datasets. CSL~\citep{murphy2019relational} contains 150 4-regular graphs that cannot be distinguished with a 1-WL isomorphism test. EXP~\citep{EXP} contains 600 pairs of non-isomorphic graphs: both 1-WL and 2-WL tests fail at differentiating these graphs. Finally, SR25~\citep{balcilar2021breaking} has 15 strongly regular graphs with 25 nodes each that cannot be discerned with a 3-WL test.  Numerical experiments show that our model 
achieves perfect accuracy on all 3 datasets while MP-GNNs fail, see Table~\ref{tab: Expressivity}. Our results are only empirical. Due to the nonlocal way in which information is passed from one layer to the other, a direct analytical comparison between the proposed neural network and the Weisfeiler-Lehman test is challenging.

\subsection{Ablation Studies}
In our ablation studies, we evaluated various choices made during the implementation of each component of the architecture. The details of these studies can be found in the appendix.
Appendix~\ref{app sec: patch extraction} focuses on the design of the patch extraction process, including the effects of the graph partition algorithm (Table~\ref{tab: patch extraction}), patch size (Figure~\ref{fig: num patch}), patch overlapping (Figure~\ref{fig: k-hop}), and other related aspects. Appendix~\ref{app sec: pe} presents the effects of two types of positional encoding, i.e., node PE and patch PE. Appendix~\ref{app sec: da} investigates the effect of data augmentation and explores the trade-off between performance and efficiency. In Appendix~\ref{section: design of MHA}, we delve into different designs of the gMHA mechanism in the Graph ViT. Additionally, we provide a complexity analysis in Appendix~\ref{sec: complexity} and discuss the limitations in Appendix~\ref{app sec: limitation}.

\section{Conclusion}\label{sec: conclusion}
In this work, we proposed a novel GNN architecture to improve standard MP-GNN limitations, particularly their low expressivity power and poor long-range dependency, and presented promising results on several benchmark graph datasets. Future work will focus on further exploring graph network architectures with the inductive biases of graph tokens and vision Transformer-like architectures in order to solve fundamental node and link prediction tasks, and possibly without the need of specialized GNN libraries like PyG \citep{pyg} or DGL \citep{zheng2020dgl} by replacing sparse linear algebra operations on graph tokens with dense operations.

\section*{Acknowledgments}
XB is supported by NUS Grant ID R-252-000-B97-133 and BH was supported in part by NUS ODPRT Grant ID A-0008067-00-00. The authors would like to express their gratitude to the reviewers for their feedback, which has improved the clarity and contribution of the paper.

\newpage
\bibliographystyle{unsrtnat}
\bibliography{main}  

\newpage
\appendix

\section{Related Work}\label{app sec: related work}

\begin{table}[!ht]
    \centering
    \scriptsize
    \caption{Comparison of different hierarchical graph models. }
    \label{tab: related work}
    \begin{tabular}{lccccc}
    \toprule
         &  GNN  
         & Transformer 
         & Graph Coarsening 
         & Local Info. 
         & Global Info.\\
         \midrule
    Coarformer~\citep{kuang2022coarformer}     
    & \cmark 
    & \cmark
    & \cmark (non-overlap, static) 
    & GNN on original graph
    & MHA on coarsen graph \\
    Exphormer~\citep{shirzad2023exphormer}
    & \cmark
    & \cmark
    & \xmark
    & GNN on original graph	
    & MHA on expander graph\\
    ANS-GT~\citep{zhang2022hierarchical_ANSGT} 
    &  \xmark
    & \cmark
    & \cmark (non-overlap, static)
    & adaptive node sampling strategy	
    & sampled nodes from the coarsened graph\\
    NAGphormer~\citep{chen2022nagphormer}
    & \xmark
    & \cmark
    & \xmark
    & MHA on multi-hop neighbour
    & --\\
    Graph MLP-Mixer (Ours) 
    & \cmark
    & \cmark
    & \cmark (overlap, dynamic)	
    & GNN on graph patches	
    & token mixer across patches\\
    \bottomrule
    \end{tabular}
\end{table}

We briefly review the hierarchical graph models~\citep{kuang2022coarformer, shirzad2023exphormer, zhang2022hierarchical_ANSGT, chen2022nagphormer} and highlight the main differences among them.

Coarformer~\citep{kuang2022coarformer} combines MPNNs and Transformers, using a GNN-based module for local information and a Transformer-based module for global information. Exphormer~\citep{shirzad2023exphormer} also employs MPNN+Transformer, using GNN and Transformer modules on the original graph and expander graph, respectively. ANS-GT~\citep{zhang2022hierarchical_ANSGT} introduces a node-sampling-based GT with hierarchical attention and graph coarsening. NAGphormer~\citep{chen2022nagphormer} treats each node as a token sequence and aggregates multi-hop information using attention-based readout.

\textbf{Main differences:}
\textbf{1) GNN/Transformer module.} Coarformer, Exphormer, SAT and ours use a hybrid MPNN+Transformer architecture while ANS-GT and NAGphormer rely solely on Transformers. However, there are notable differences between these approaches as we do not use any Transformer but rather MLP as our backbone. Besides, our MPNN operates on small graph patches instead of the entire graph as Coarformer and Exphormer. Furthermore, SAT and our architecture are sequential, while Coarformer and Exphormer combine MP-GNNs and GT in parallel.

\textbf{2) Graph coarsening module.} Coarformer, ANS-GT and SAT use a graph coarsening mechanism. These methods perform graph coarsening as a pre-processing step and generate static and non-overlapping graph patches. In contrast, we perform graph coarsening with a stochastic version of Metis on-the-fly, generating dynamic and overlapping graph patches.

\textbf{3) Graph embedding module.} The ways to capture local and global information are different as stated in the above reviews and the above summary Table~\ref{tab: related work}.

In summary, although these aforementioned hierarchical graph models share similarities with our model, the major difference between these models and our work is that we do not use Graph Transformer (GT) as the backbone architecture but an alternative architecture based on ViT/MLP-Mixer and graphs. We believe moving from GT to Graph ViT/MLP-Mixer as a new backbone/high-level architecture has the potential to open a new line of work for GNN design (by enhancing the building blocks of the proposed architecture such as graph clustering, graph embedding, mixer layer, positional encoding, (pre-)training, etc).

\section{Datasets Description}\label{app_sec: Datasets description}
\begin{table}[!ht]
\caption{Summary statistics of datasets used in this study}
\footnotesize
    \centering
    \begin{tabular}{lccccccc}
    \toprule
    Dataset     &  \#Graphs & \#Nodes & Avg. \#Nodes & Avg. \#Edges & Task & Metric\\
    \midrule
    CSL & 150 & 41 & 41 & 164 & 10-class classif. & Accuracy\\
    EXP & 1,200 & 32-64 & 44.4 & 110.2 & 2-class classif. & Accuracy\\
    {SR25} & {15} & {25} & {25} &{300} & {15-class classif.} & {Accuracy} \\
    \midrule
         ZINC & 12,000 & 9-37 & 23.2 & 24.9 & regression & MAE \\
         MNIST & 70,000 & 40-75 & 70.6 & 684.4 & 10-class classif. & Accuracy \\
         CIFAR10 & 60,000 & 85-150 & 117.6 & 1129.7 & 10-class classif. & Accuracy\\
    \midrule
        MolTOX21 & 7,831 & 1-132 & 18.57 & 38.6 & 12-task classif. & ROCAUC\\
         MolHIV & 41,127 & 2-222 & 25.5 & 54.9 & binary classif. & ROCAUC \\
    \midrule
         {Peptides-func} 
         & {15,535} 
         & {8-444} 
         & {150.9} 
         & {307.3} 
         & {10-class classif.} 
         & {Avg. Precision (AP)} \\
         {Peptides-struct} 
         & {15,535} 
         & {8-444} 
         & {150.9} 
         & {307.3} 
         & {regression} 
         & {MAE} \\
    \midrule
         TreeNeighbourMatch (r=2)
         &96 &7 &7 &6 & 4-class classif. & Accuracy \\
         TreeNeighbourMatch (r=3)
         & 32,000 & 15 & 15 & 14 & 8-class classif. & Accuracy\\
         TreeNeighbourMatch (r=4)
         & 64,000 & 31 & 31 & 30 & 16-class classif. & Accuracy\\
         TreeNeighbourMatch (r=5)
         & 128,000 & 63 & 63 & 62 & 32-class classif. & Accuracy\\
         TreeNeighbourMatch (r=6)
         & 256,000 & 127 & 127 & 126 & 64-class classif. & Accuracy\\
         TreeNeighbourMatch (r=7)
         & 512,000 & 255 & 255 & 254 & 128-class classif. & Accuracy\\
         TreeNeighbourMatch (r=8)
         & 640,000 & 511 & 511 & 510 & 256-class classif. & Accuracy\\
    \bottomrule
    \end{tabular}
    \label{tab: datasets}
\end{table}

We evaluate our Graph MLP-Mixer on a wide range of graph benchmarks. See summary statistics of datasets in Table~\ref{tab: datasets}.

\textbf{CSL}~\citep{murphy2019relational} is a synthetic dataset to test the expressivity of GNNs.  CSL has 150 graphs divided into 10 isomorphism classes. Each CSL graph is a 4-regular graph with edges connected to form a cycle and containing skip-links between nodes.  The goal of the task is to classify them into corresponding isomorphism classes. 

\textbf{EXP}~\citep{EXP} contains 600 pairs of 1\&2-WL failed graphs. The goal is to map these graphs into two classes.

\textbf{SR25}~\citep{balcilar2021breaking} has 15 strongly regular graphs (3-WL failed) with 25 nodes each. SR25 is translated to a 15 way classification problem with the goal of mapping each graph into a different class.

\textbf{ZINC}~\citep{dwivedi2020benchmarking} is a subset (12K) of molecular graphs (250K) from a free database of commercially-available compounds~\citep{ZINC}. These molecular graphs are between 9 and 37 nodes large. Each node represents a heavy atom (28 possible atom types) and each edge represents a bond (3 possible types). The task is to regress a molecular property known as the constrained solubility. The dataset comes with a predefined 10K/1K/1K train/validation/test split.

\textbf{MNIST and CIFAR10}~\citep{dwivedi2020benchmarking} are derived from classical image classification datasets by constructing an 8 nearest-neighbor graph of SLIC superpixels for each image. The resultant graphs are of sizes 40-75 nodes for MNIST and 85-150 nodes for CIFAR10. The 10-class classification tasks and standard dataset splits follow the original image classification datasets, i.e., for MNIST 55K/5K/10K and for CIFAR10 45K/5K/10K train/validation/test graphs.  These datasets are sanity-checks, as we expect most GNNs to perform close to 100\% for MNIST and well enough for CIFAR10.

\textbf{MolTOX21 and MolHIV}~\citep{hu2020open} are molecular property prediction datasets adopted from the MoleculeNet~\citep{MoleculeNet}. All the molecules are pre-processed using RDKit~\citep{landrum2006rdkit}. Each graph represents a molecule, where nodes are atoms, and edges are chemical bonds. Input node features are 9-dimensional, containing atomic number and chirality, as well as other additional atom features such as formal charge and whether the atom is in the ring or not. The datasets come with a predefined scaffold splits based on their two-dimensional structural frameworks, i.e. for MolTOX21 6K/0.78K/0.78K and for MolHIV 32K/4K/4K train/validation/test.

{\textbf{Peptides-func and Peptides-struct}~\citep{dwivedi2022long}
are derived from 15,535 peptides with a total of 2.3 million nodes retrieved from SAT-Pdb~\citep{singh2016satpdb}. Both datasets use the same set of graphs but differ in their prediction tasks. These graphs are constructed in such a way that requires long-range interactions (LRI) reasoning to achieve strong performance in a given task. In concrete terms, they are larger graphs: on average 150.94 nodes per graph, and on average 56.99 graph diameter. Thus, they are better suited to benchmarking of graph Transformers or other expressive GNNs that are intended to capture LRI.}

\textbf{TreeNeighbourMatch} is a synthetic dataset proposed by~\citet{alon2020bottleneck} to highlight the inherent problem of over-squashing in GNNs. It is designed to simulate the exponentially-growing receptive field while allowing us to control the problem radius $r$, and thus control the intensity of over-squashing. Specifically, each graph is a binary tree of depth $depth$ (a.k.a. problem radius $r$). The goal is to predict a label for the target node, where the correct answer lies in one of the leave nodes. Therefore, the TreeNeighbourMatch problem requires information to be propagated from all leave nodes to the target node before predicting the label, causing the issue of over-squashing at the target node.

\begin{figure}[!ht]
\centering     
\subfigure[ZINC]{\includegraphics[scale=0.3]{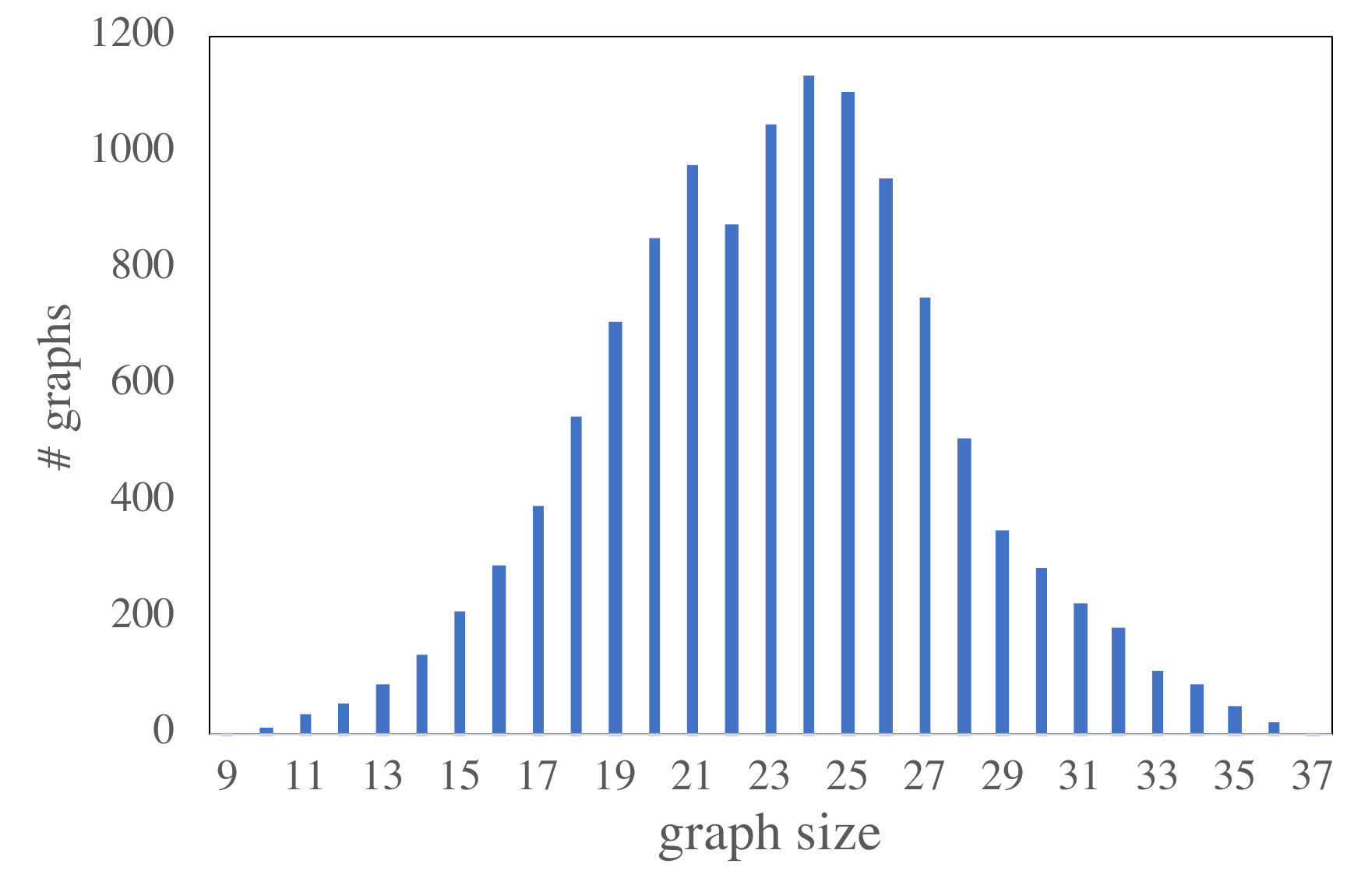}}
\subfigure[MNIST]{\includegraphics[scale=0.3]{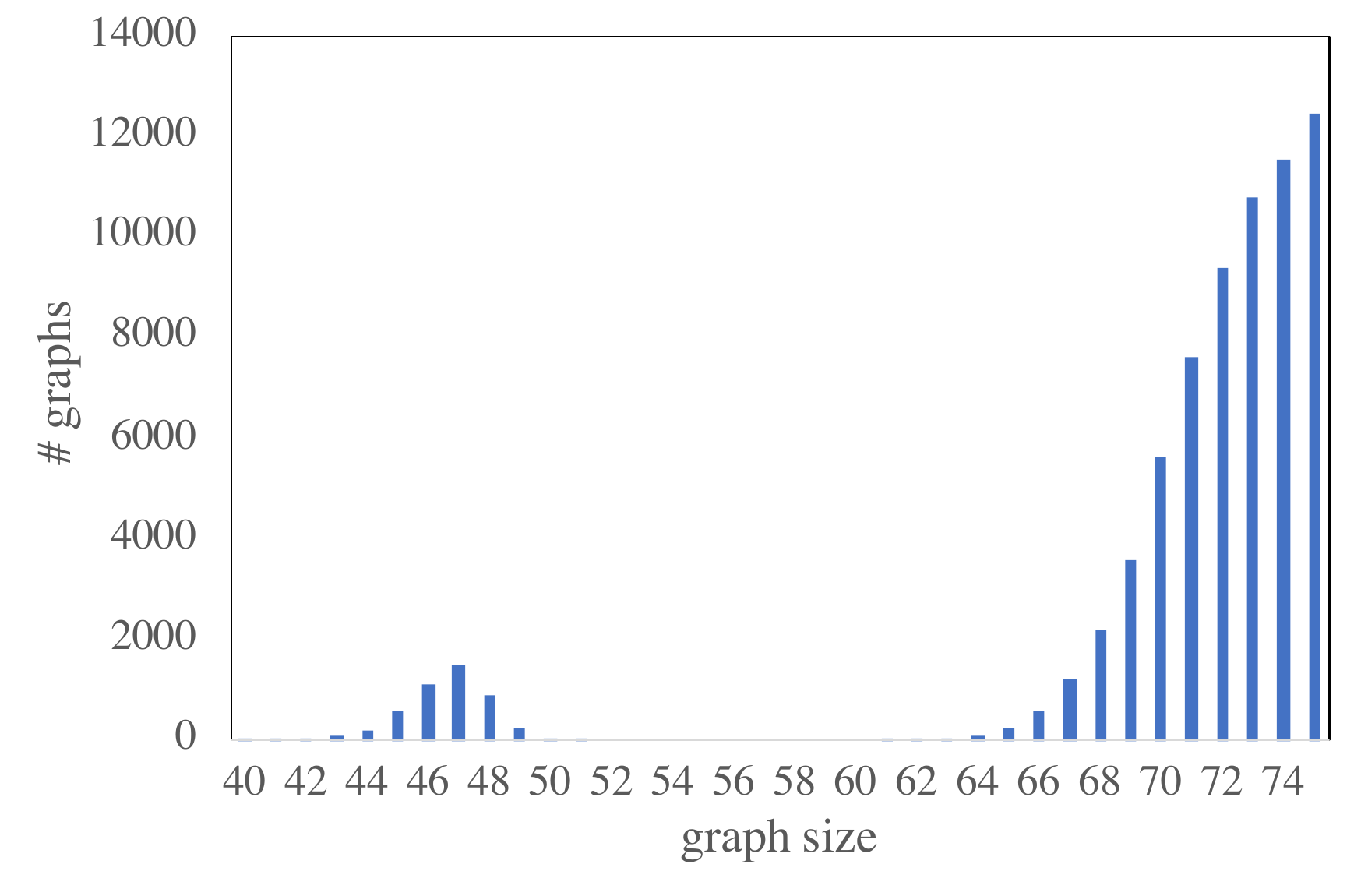}}
\subfigure[CIFAR10]{\includegraphics[scale=0.3]{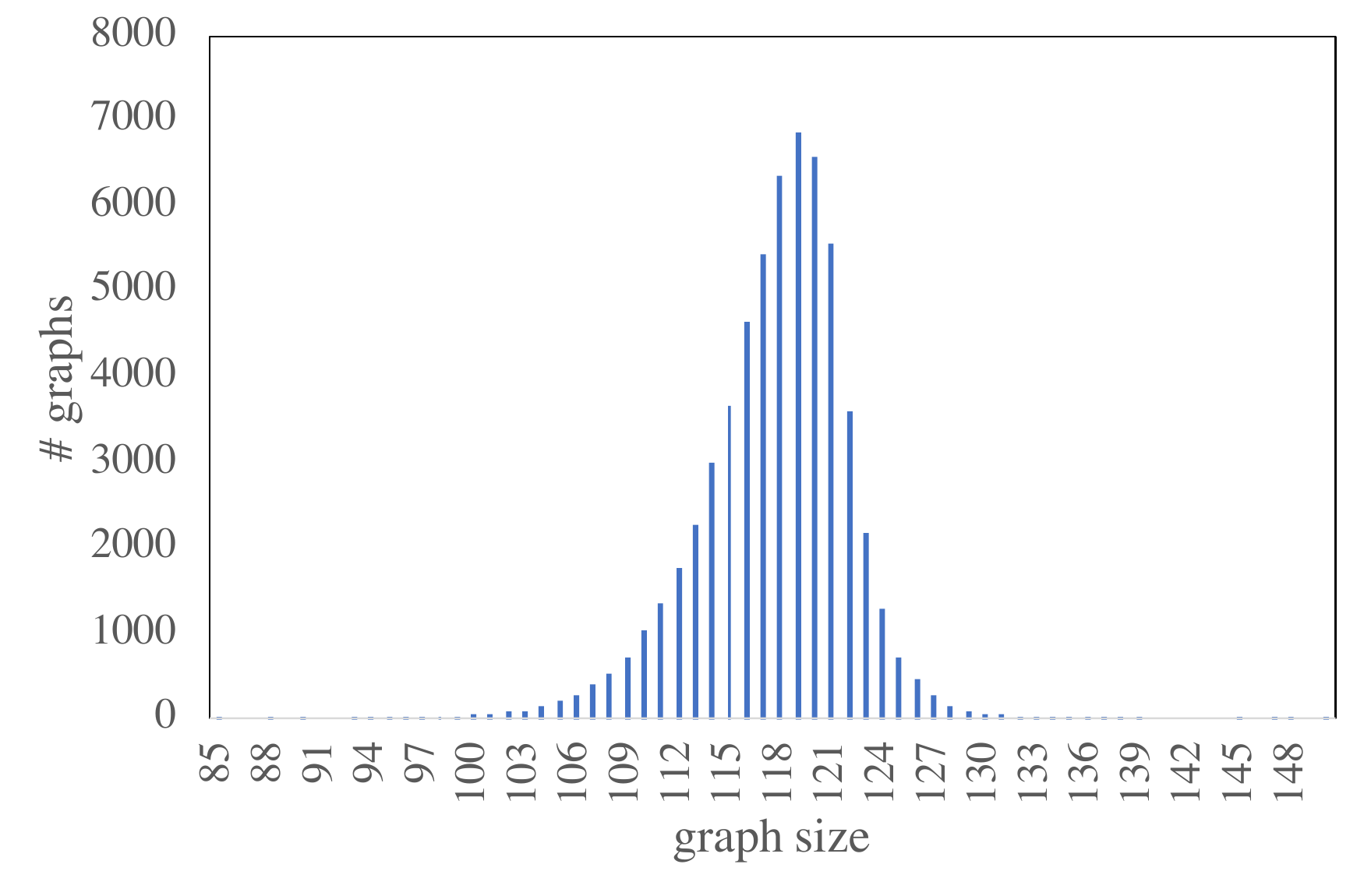}}
\subfigure[MolTOX21]{\includegraphics[scale=0.3]{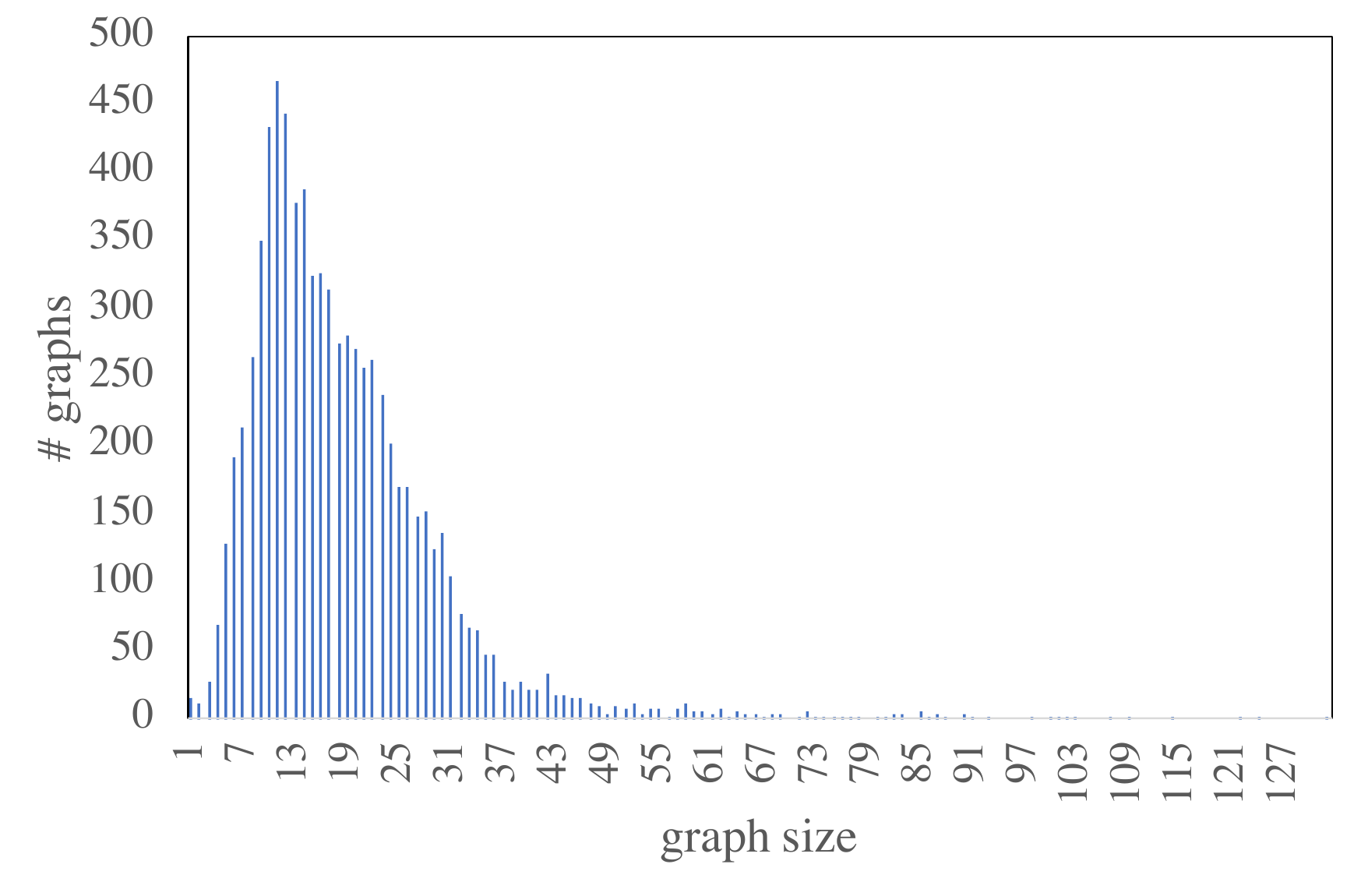}}
\subfigure[MolHIV]{\includegraphics[scale=0.3]{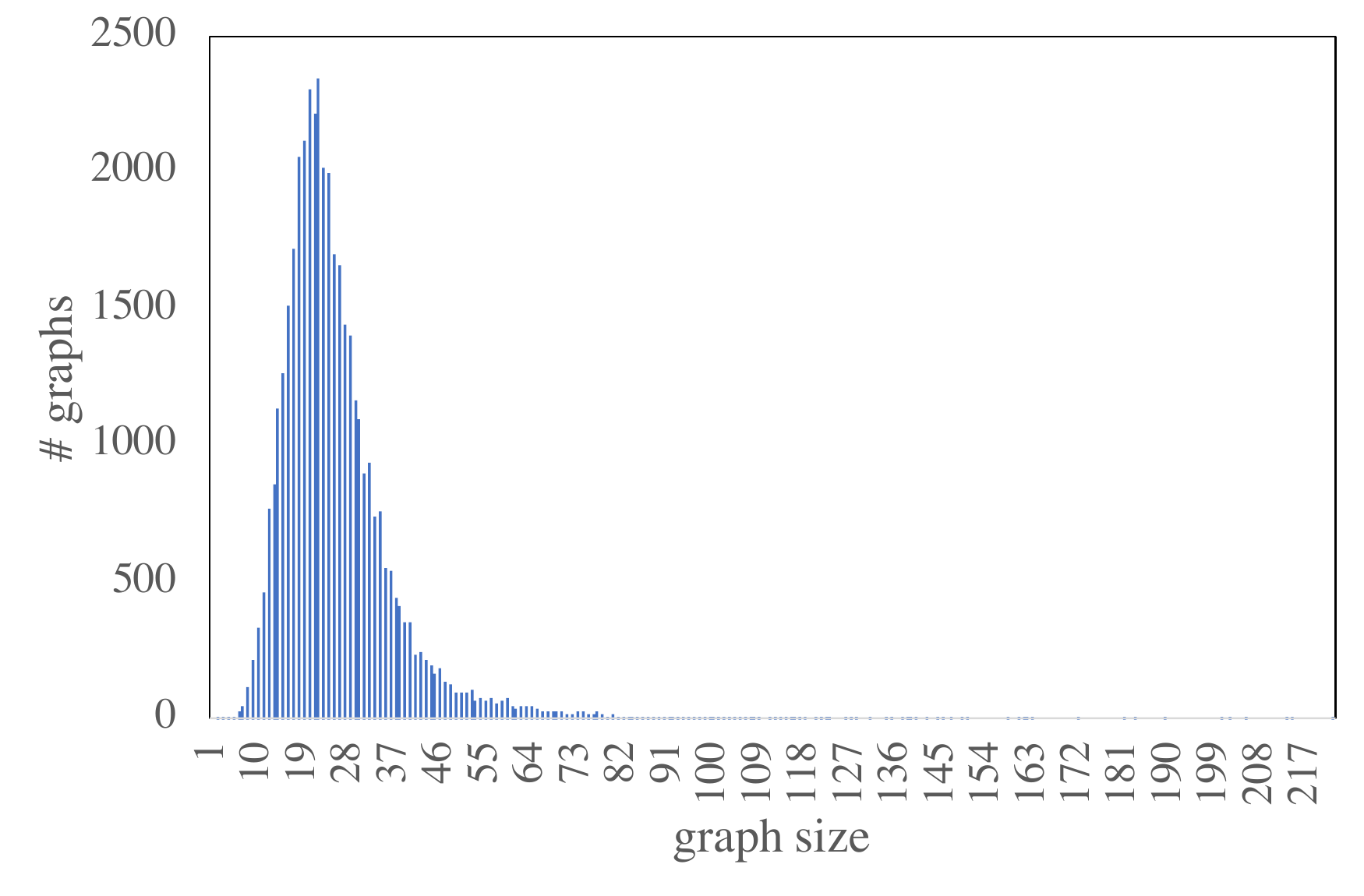}}
\subfigure[Peptides-func/struct]{\includegraphics[scale=0.3]{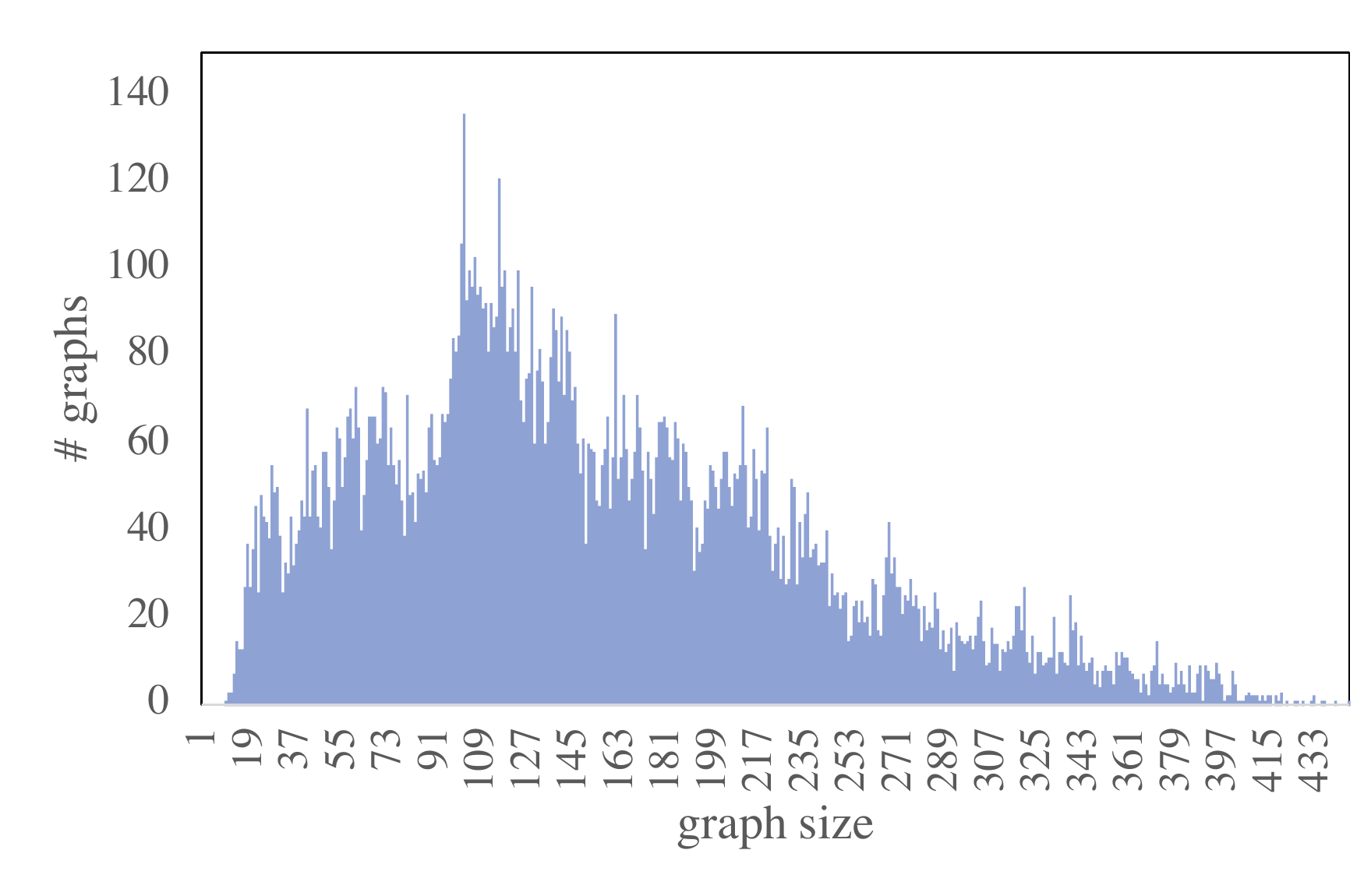}}
\caption{Distributions of the graph sizes.}
\label{fig: Distributions of the graph sizes}
\end{figure}

\textbf{Distributions of the graph sizes.} We plot of the distributions of the graph sizes (i.e., the number of nodes in each data sample) of these datasets in Figure~\ref{fig: Distributions of the graph sizes}.

\begin{table}[!ht]
\caption{Summary statistics of graph patches.}
\footnotesize
    \centering
    \begin{tabular}{lm{4em}m{4em}m{4em}m{4em}m{4em}m{4em}m{4em}}
    \toprule
         \multirow{2}{*}{Dataset} & \multirow{2}{*}{\# Patch}
         & \multicolumn{3}{c}{\# Node} & \multicolumn{3}{c}{Diameter} \\
        \cmidrule(lr){3-5}\cmidrule(lr){6-8}
        &  & Mean & Min & Max & Mean & Min & Max \\
         \midrule
         CSL & 32 & 5.80 & 5 & 8 & 2.28 & 2 & 3 \\
         EXP & 32 & 4.07 & 2 & 11 & 2.31 & 1 & 5 \\
         SR25 & 32 & 13.00 & 13 & 13 & 2.00 & 2 & 2\\
         \midrule
         ZINC  
         & 32 
         & 3.15 & 2 & 7 
         & 1.82 & 1 & 3 \\
         MNIST 
         & 32 
         & 14.36 & 9 & 28 
         & 2.85 & 2 & 5 \\
         CIFAR10 
         & 32 
         &17.20 &10 &35
         &3.07 &2 &7 \\
         \midrule
         MolTOX21 
         & 32 
         & 3.15 & 1 & 10 
         & 1.80 & 0 & 6 \\
         MolHIV 
         & 32 
         &3.27 &1 &13
         &1.87 &0 &8 \\
         \midrule
         {Peptides-func}
         &32
         &7.08 &1 &20
         &4.15 &0 &14 \\
         {Peptides-struct}
         &32
         &7.08 &1 &20
         &4.15 &0 &14 \\
         \midrule
         TreeNeighbourMatch(r=2) & 8  & 1.86 &1 &3 &0.86 &0 &2\\
         TreeNeighbourMatch(r=3) & 32 &1.93 &1 &3 &0.93 &0 &2\\
         TreeNeighbourMatch(r=4) & 32 &1.97 &1 &3 &0.97 &0 &2\\
         TreeNeighbourMatch(r=5) & 32 &3.28 &1 &5  & 2.25 &0 &3\\
         TreeNeighbourMatch(r=6) & 32 &5.34 &3 &8 &3.31 &2 &5\\
         TreeNeighbourMatch(r=7) & 32 &9.19 &7 &14 &4.33 &4 &5\\
         TreeNeighbourMatch(r=8) & 32 &17.03 &15 &23 &6.17 &6 &8\\
         \bottomrule
    \end{tabular}
    \label{tab: graph patch statistics}
\end{table}

\textbf{Patch size and diameter.}
We set the number of patches to 32 by default. Summary statistics of graph patches are presented in Table~\ref{tab: graph patch statistics}.

\section{Experiment Details}\label{app: experiment}

We implement out model using PyTorch~\citep{pytorch} and PyG~\citep{pyg}. We ran our experiments on NVIDIA RTX A5000 GPUs.
We run each experiment with 4 different seeds, reporting the averaged results at the epoch achieving the best validation metric. For optimization, we use Adam~\citep{kingma2014adam} optimizer, with the default settings of $\beta_1=0.9$, $\beta_2=0.999$, and $\epsilon=1e^{-8}$. 
We observe large fluctuations in the validation metric with the common Adam optimizer on the OGB datasets (i.e., MolHIV and MolTOX21), as also observed in~\citep{sun, zhang2021nested, chen2019equivalence}. 
We consider following the practice of SUN~\citep{sun} by employing the ASAM optimizer~\citep{kwon2021asam} to reduce such fluctuations. We use the same hyperparameter with batch size of 32 and learning rate of 0.01 without further tuning.

\textbf{Simulation Datasets.}
For CSL and EXP, we run the 5-fold cross validation with stratified sampling to ensure class distribution remains the same across the splits~\cite{dwivedi2020benchmarking, zhang2021nested}. For SR25 dataset, we follow the evaluation process in~\citep{zhao2021stars, feng2022powerful} that directly train and validate the model on the whole dataset and report the best performance.

\textbf{Real-World Datasets.}
For benchmarking datasets from~\citet{dwivedi2020benchmarking}, we followed the most commonly used parameter budgets:  up to 500k parameters for ZINC; For MolTOX21 and MolHIV from OGB~\citep{hu2020open}, there is no upper limit on the number of parameters. For peptides-func and peptides-struct from LRGB~\citep{dwivedi2022long}, we followed the parameter budget $\sim$500k. All real world evaluated benchmarks define a standard train/validation/test dataset split.

\textbf{Baselines.} 
We use GCN~\citep{kipf2017semi}, GatedGCN~\citep{bresson2017gatedgcn}, GINE~\citep{hu2019gine} and Graph Transformer~\citep{dwivedi2021generalization} as our baseline models, which also server as the base patch encoder of Graph MLP-Mixer. The hidden size is set to 128 and the number of layers is set to 4 by default. For TreeNeighbourMatch datasets, we follow the experimental protocol introduced in~\citep{alon2020bottleneck}, that is, for TreeNeighbourMatch dataset with problem radias $r=\textit{depth}$, we implemented a network with $r+1$ graph layers to allow an additional nonlinearity after the information from the leaves reaches the target node.

\textbf{Graph MLP-Mixer.}
The hidden size is set to 128, and the number of GNN layers and Mixer layers is set to 4. Except that for LRGB datasets, we reduce the number of Mixer layers to 2 to fulfill the parameter budget $\sim$500k.


\textbf{SOTA models.}
In Table~\ref{tab: sota} and Table~\ref{app tab: lrgb}, results are referenced directly from literature if available, otherwise are reproduced using authors' official code.
To enable a fair comparison of speed/memory complexity (Table~\ref{app tab: lrgb}), we set the batch size to 128 all the SOTA models and ours and reduce the batch size by half if OOM until the model and batch data can be fit into the memory. Besides, all experiments are run on the same machine.

\textbf{Positional Encodings.}
As the most appropriate choice of node positional encoding (NodePE) is dataset and task dependent, we follow the practice of~\citet{rampavsek2022recipe, dwivedi2022long}, see Table~\ref{tab: pe}. We have already augmented all the base models (GCN, GatedGCN, GINE and GraphTrans) in Table~\ref{tab: performance} with the same type of NodePE as Graph MLP-Mixer to ensure a fair comparison.
\begin{table}[!ht]
    \centering
    \scriptsize
    \caption{Summary statistics of positional encoding (PE).}
    \begin{tabular}{lcccccccccc}
    \toprule
          &  CSL & EXP& SR25 & ZINC & MNIST & CIFAR10 &MolTOX21 & MolHIV & Peptides-fun & Peptides-struct\\
         \midrule
         NodePE 
         & RWSE-8 & RWSE-8 & LapPE-8 
         &RWSE-20 & LapPE-8 & LapPE-8
         & -- & --
         & RWSE-16 & RWSE-16 \\
         PatchPE 
         & RWSE-8 & RWSE-8 & RWSE-8 & RWSE-8
         & RWSE-8 & RWSE-8
         & -- & --
         & RWSE-8 & RWSE-8 \\
         \bottomrule
    \end{tabular}
    \label{tab: pe}
\end{table}

\section{Studies on Patch Extraction Module}\label{app sec: patch extraction}

\subsection{Effect of Patch Extraction}
\begin{table}[!ht]
    \caption{Effect of patch extraction: \xmark means no patch extraction and \cmark means uses patch extraction.
    \label{tab: patch extraction}}
    \centering
    \small
    \begin{tabular}{lccc}
    \toprule
         Model &  Patch Extraction & ZINC (MAE$\downarrow$) & Peptides-func (AP$\uparrow$)\\
        \midrule
         \multirow{2}{*}{GCN-MLP-Mixer}
         & \xmark & 0.2495 ± 0.0040 & 0.6341 ± 0.0139 \\
         & \cmark & 0.1347 ± 0.0020 & 0.6832 ± 0.0061 \\
         \midrule
         \multirow{2}{*}{GatedGCN-MLP-Mixer}
         & \xmark & 0.2521 ± 0.0084 & 0.6230 ± 0.0110\\
         & \cmark & 0.1244 ± 0.0053  & 0.6932 ± 0.0017\\
         \midrule
         \multirow{2}{*}{GINE-MLP-Mixer}
         & \xmark & 0.2558 ± 0.0059 & 0.6350 ± 0.0038\\
         & \cmark & 0.0733 ± 0.0014 & 0.6970 ± 0.0080\\
         \midrule
         \multirow{2}{*}{GraphTrans-MLP-Mixer}
         & \xmark & 0.2538 ± 0.0067 & 0.6224 ± 0.0112\\
         & \cmark & 0.0773 ± 0.0030 & 0.6858 ± 0.0062\\
         \bottomrule
    \end{tabular}
\end{table}
We conducted an experiment where we ran the Graph MLP-Mixer without the patch extraction process, treating each individual node as a patch. The results of this experiment are presented in Table~\ref{tab: patch extraction}. The patch extraction process is critical. We believe that the patch extraction process, which includes Metis partition and 1-hop extension, helps to capture important local information about the graph structure.

\subsection{Effect of Graph Partition Algorithm}
\begin{table}[!ht]
\caption{Comparison of METIS and random graph partition algorithm.}
    \centering
    \footnotesize
    \begin{tabular}{lcccc}
    \toprule
         \multirow{2}{*}{Model}
         & \multicolumn{2}{c}{ZINC (MAE $\downarrow$)} 
         & \multicolumn{2}{c}{Peptides-struct (MAE $\downarrow$)} \\
         \cmidrule(lr){2-3}\cmidrule(lr){4-5}
         & METIS & Random & METIS & Random\\
         \midrule
    GCN-MLP-Mixer     
    & 0.1347 ± 0.0020 & 0.1435 ± 0.0122
    & 0.2486 ± 0.0041 & 0.2565 ± 0.0031\\
    GatedGCN-MLP-Mixer
    & 0.1244 ± 0.0053 & 0.1284 ± 0.0074
    & 0.2508 ± 0.0007 & 0.2539 ± 0.0012\\
    GINE-MLP-Mixer
    & 0.0733 ± 0.0014 & 0.0708 ± 0.0020
    & 0.2494 ± 0.0007 & 0.2559 ± 0.0012\\
    GraphTrans-MLP-Mixer
    & 0.0773 ± 0.0030 & 0.0767 ± 0.0019
    & 0.2480 ± 0.0013 & 0.2574 ± 0.0025\\
    \bottomrule
    \end{tabular}
    \label{tab:metis vs random graph partition}
\end{table}

Graph partitioning algorithms have been studied for decades~\citep{bulucc2016recent} given their importance in identifying meaningful clusters. Mathematically, graph partitioning is known to be NP-hard~\citep{chung1997spectral}. Approximations are thus required. A graph clustering algorithm with one of the best trade-off accuracy and speed is METIS~\citep{karypis1998metis}, which partitions a graph into a pre-defined number of clusters/patches such that the number of within-cluster links is much higher than between-cluster links in order to better capture good community structure. For these fine properties, we select METIS as our graph patch extraction algorithm.

We provide the ablation study for how many benefits the METIS can provide against random graph partitioning. For random graph partition, nodes are randomly assigned to a pre-defined number of patches. We apply data augmentation as described in Section~\ref{subsec: data augmentation} to both algorithms. Table~\ref{tab:metis vs random graph partition} shows that using METIS as the graph partition algorithm consistently gives better performance than random node partition, especially on graphs with more nodes and edges (such as Peptides-func), which corresponds to our intuition that nodes and edges composing a patch should share similar semantic or information. 
Nevertheless, it is interesting to see that random graph partitioning is still able to achieve reasonable results, which shows that the performance of the model is not solely supported by the quality of the patches.

\subsection{Effect of Number of Patches}

\begin{figure}[!ht]
\centering     
\subfigure{\includegraphics[scale=0.35]{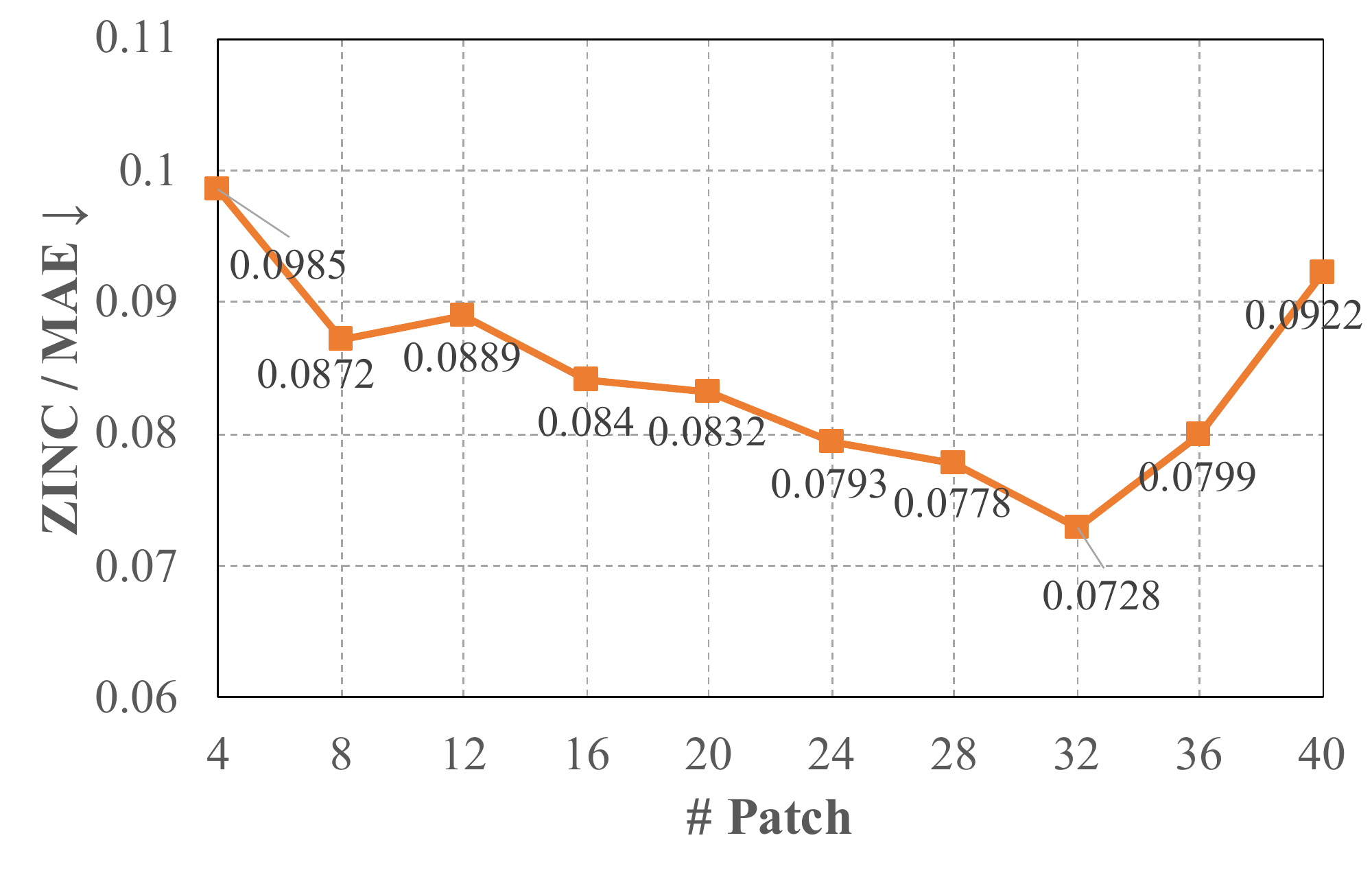}}
\hspace{8mm}
\subfigure{\includegraphics[scale=0.35]{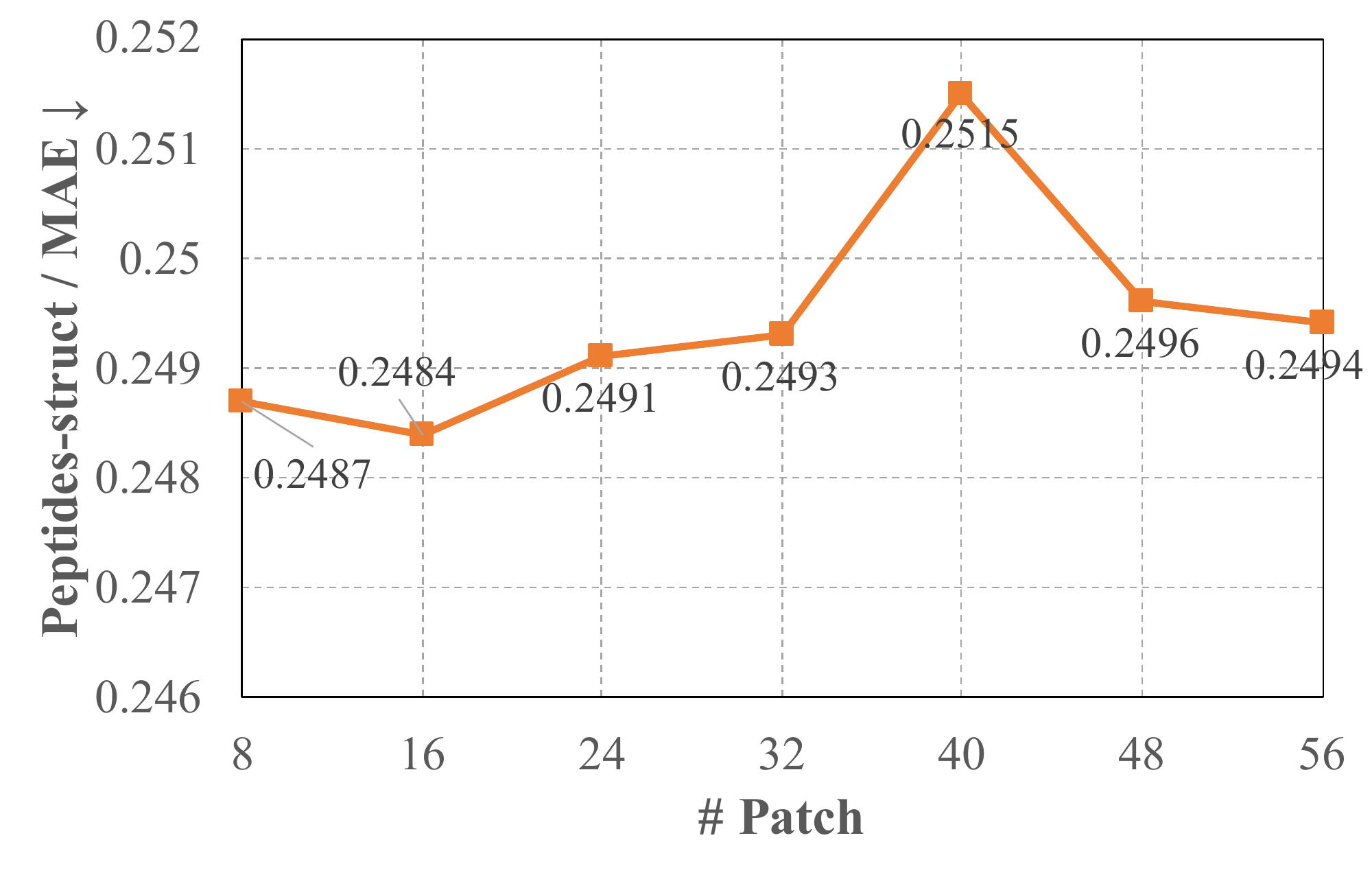}}
\caption{Effect of the number of patches.}
\label{fig: num patch}
\end{figure}

We observe in Figure~\ref{fig: num patch} when increasing the number of graph patches (\# Patch), performance increases first and then flattens out (with small fluctuations) when \#Patch=24. We set the number of patches to 32 by default.

\subsection{Effect of Patch Overlapping}
\begin{figure}[!ht]
\centering     
\subfigure{\includegraphics[scale=0.45]{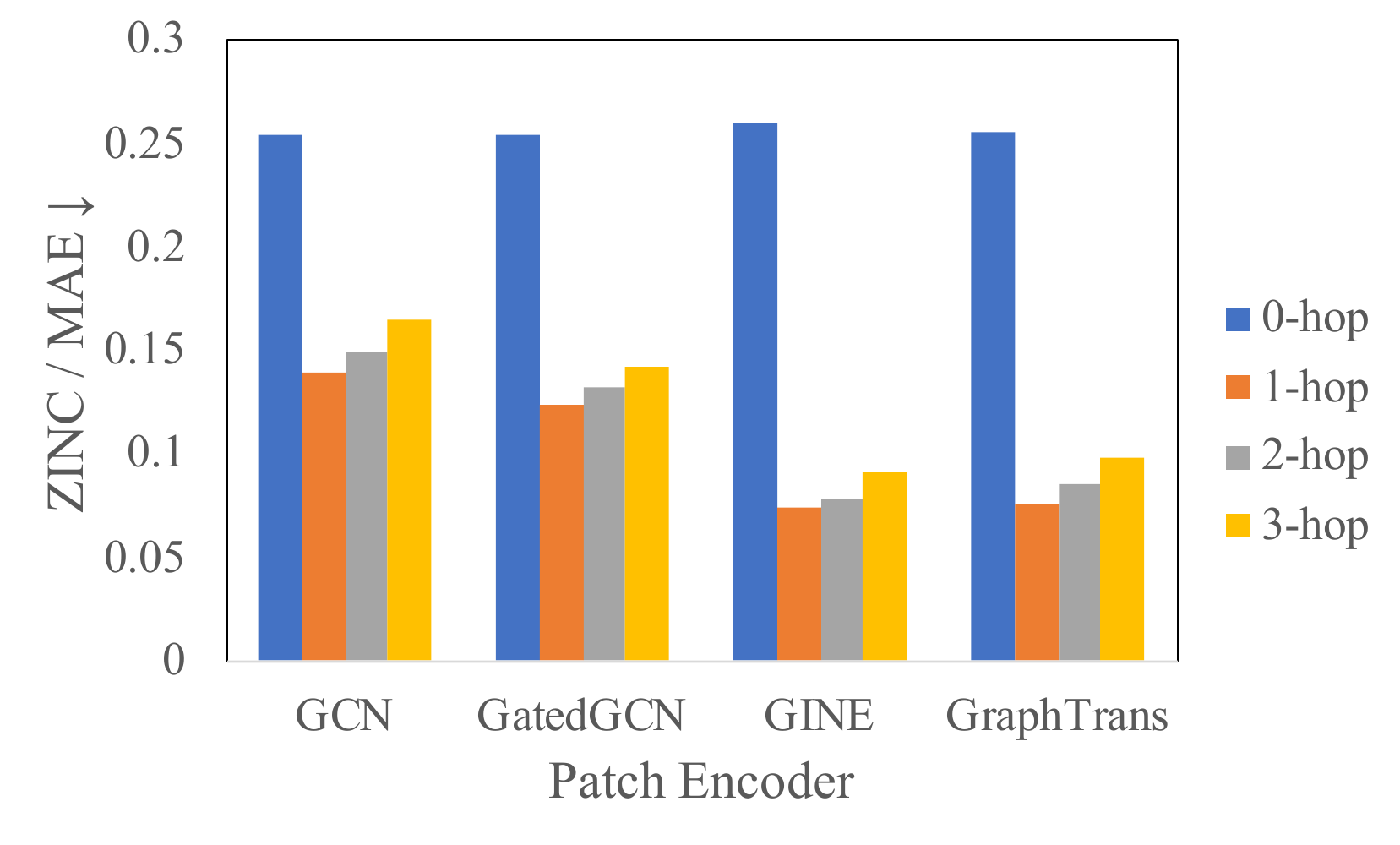}}
\subfigure{\includegraphics[scale=0.45]{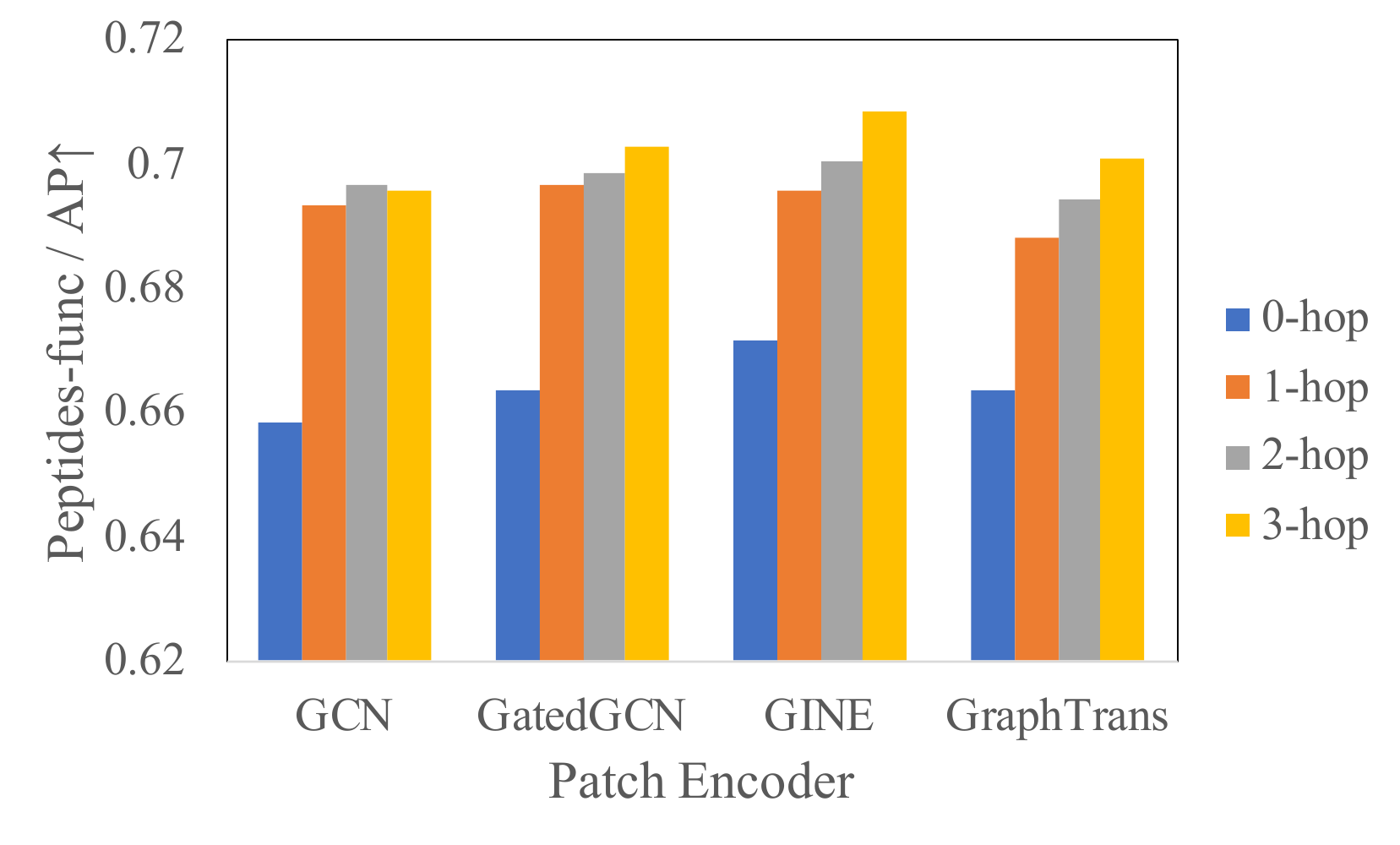}}
\caption{Effect of the patch overlapping with $k$-hop extension.}
\label{fig: k-hop}
\end{figure}
In Figure~\ref{fig: k-hop}, we observe a clear performance increase when graph patches are overlapping with each other (0-hop vs 1-hop), which is consistent with our intuition that extracting non-overlapping patches implies losing important edge information. We further expand graph patches to their $k$-hop neighbourhood. Performance increases first and then flattens out or begins to decrease when $k=2$ for ZINC and $k=3$ for Peptides-func. We set $k=1$ by default.

\subsection{Patch Size and WL Expressivity}

\begin{table}[!ht]
  \caption{Accuracy on EXP with different patch sizes $P$, averaging over 4 runs with 4 different seeds}
  \label{tab: exp}
    \centering
    \small
    \begin{tabular}{lccccc}
    \toprule
    Model & P=2 &P=4 & P=8 & P=16 & P=32 \\
    \midrule
     {GCN-MLP-Mixer}
     & 57.54 ± 3.87 
     & 99.44 ± 0.59 &99.69 ± 0.98 & 100.00 ± 0.00 & 100.00 ± 0.00  \\
     {GatedGCN-MLP-Mixer}
     & 67.65 ± 2.01 
     & 99.77 ± 0.37 & 100.00 ± 0.00 & 100.00 ± 0.00 & 100.00 ± 0.00  \\
     {GINE-MLP-Mixer}
     & 57.75 ± 3.80 
     & 99.58 ± 0.45 
     & 100.00 ± 0.00 & 100.00 ± 0.00 & 100.00 ± 0.00  \\
     {GraphTrans-MLP-Mixer}& 73.79 ± 1.52 & 96.77 ± 8.43 & 100.00 ± 0.00& 100.00 ± 0.00 & 100.00 ± 0.00  \\
     \bottomrule
    \end{tabular}
    \end{table}
    
We evaluated the 2-WL and 3-WL expressivity on the benchmark datasets available to us, which indeed have small graphs. As we do not have access to 2-WL/3-WL datasets with larger graph sizes, we studied the impact of performance with a smaller number of patches in Table~\ref{tab: exp}. As expected, expressivity increases when the number of patches increases as well. Given these experiential results, we also suppose that for larger graphs, we would need to increase the number of patches to maintain expressivity.

\section{Studies on Positional Encoding}\label{app sec: pe}


\subsection{Effect of Positional Encoding}

\begin{table}[!ht]
    \centering
    \small
    \caption{Effect of positional encoding. We study the effects of node PE and patch PE by removing one of them in turn from our model while keeping the other components unchanged.}
    \label{tab: node PE and patch PE}
    \begin{tabular}{llccccc}
    \toprule
    Dataset &Method&  GCN-MLP-Mixer & Gated-MLP-Mixer & GINE-MLP-Mixer & GraphTrans-MLP-Mixer\\
         \midrule
    \multirow{4}{*}{ZINC}
    & Full     
    & 0.1347 ± 0.0020	
    & 0.1244 ± 0.0053 
    & 0.0733 ± 0.0014 
    & 0.0773 ± 0.0030\\
    & - NodePE
    & 0.1944 ± 0.0061	
    & 0.1775 ± 0.0031
    & 0.1225 ± 0.0070	
    & 0.1393 ± 0.0122\\
    & - PatchPE
    &0.1414 ± 0.0058	
    &0.1250 ± 0.0026
    &0.0746 ± 0.0010	
    &0.0778 ± 0.0029\\
    & - Both
    &0.2207 ± 0.0072
    &0.1883 ± 0.0096
    &0.1160 ± 0.0023
    &0.1700 ± 0.0064\\
    \midrule
    \multirow{4}{*}{Peptides-func}
    & Full     
    & 0.6832 ± 0.0061
    & 0.6932 ± 0.0017
    & 0.6970 ± 0.0080		
    & 0.6858 ± 0.0062\\
    & - NodePE
    & 0.6688 ± 0.0039
    & 0.6864 ± 0.0080
    & 0.6868 ± 0.0034		
    & 0.6763 ± 0.0030\\
    & - PatchPE
    & 0.6871 ± 0.0055
    & 0.6934 ± 0.0055
    & 0.6933 ± 0.0104		
    & 0.6882 ± 0.0076\\
    & - Both
    & 0.6760 ± 0.0078
    & 0.6847 ± 0.0034
    & 0.6756 ± 0.0070		
    & 0.6783 ± 0.0088\\
    \bottomrule
    \end{tabular}
\end{table}
It was proved in~\citep{murphy2019relational, Loukas2020What} that unique and permutation-invariant positional encoding (PE) increases the representation power of any MP-GNN, i.e. PE leads to GNNs strictly more powerful than the 1-WL test. PE is thus important from a theoretical point of view but, unfortunately, theory does not provide any guidance on the choice of PE for a given graph dataset and task. Consequently, the choice of PE is so far arbitrary and is selected by trial-and-error experiments such as~\citep{rampavsek2022recipe, lim2022sign} to cite the most recent PE-based GNNs. 

Our experiments show that PE may or not be useful, see Table~\ref{tab: node PE and patch PE}. Thus, PE increases the expressivity power of GNNs but not necessarily their generalization performance. In other words, they improve over-fitting but not necessarily generalization. In conclusion, PE is certainly useful to improve the quality of GNN prediction given the theory and the increased number of published works on this topic, but more mathematical progress is needed to identify more relevant choices and provides consistent result improvement.

\subsection{Positional Encoding and Patch Size}
\begin{table}[!ht]
    \centering
    \small
    \caption{Ablation with combining effects of PE and patch size on ZINC.}
    \label{tab: pe and patch size (zinc)}
    \begin{tabular}{lcccc}    
    \toprule
    Patch Size & 2 & 4 & 16 & 32 
    \\
    \midrule
    Full   
    & 0.0983 ± 0.0042 
    & 0.1011 ± 0.0103
    & 0.0799 ± 0.0037
    & 0.0743 ± 0.0049 
    \\
    
     - Node PE 
    & 0.1589 ± 0.0056
    & 0.1414 ± 0.0061
    & 0.1307 ± 0.0107
    & 0.1154 ± 0.0032 
    \\
     
     - Patch PE
    & 0.1081 ± 0.0007
    & 0.1076 ± 0.0110
    & 0.0840 ± 0.0035
    & 0.0744 ± 0.0037 
    \\
     
     - Both 
    & 0.1677 ± 0.0045
    & 0.1532 ± 0.0051
    & 0.1284 ± 0.0018
    & 0.1187 ± 0.0050 
    \\
     \bottomrule
    \end{tabular}
\end{table}

\begin{table}[!ht]
    \centering
    \small
    \caption{Ablation with combining effects of PE and patch size on Peptide-func.}
    \label{tab: pe and patch size (peptide)}
    \begin{tabular}{lccccc}
    \toprule
    Patch Size & 2 & 4 & 16 & 32 & 64 
    \\
    \midrule
    Full     
    & 0.6578 ± 0.0063 
    & 0.6675 ± 0.0037 
    & 0.6855 ± 0.0039
    & 0.6939 ± 0.0034 
    & 0.6944 ± 0.0074
    \\
    
     - Node PE 
     & 0.6613 ± 0.0063
     & 0.6708 ± 0.0065
     & 0.6864 ± 0.0069
     & 0.6873 ± 0.0033 
     & 0.6789 ± 0.0047
     \\
     
     - Patch PE
     & 0.6594 ± 0.0059
     & 0.6724 ± 0.0051
     & 0.6937 ± 0.0068
     & 0.6939 ± 0.0062
     & 0.6865 ± 0.0061
     \\
     
     - Both 
     & 0.6562 ± 0.0057
     & 0.6739 ± 0.0038
     & 0.6879 ± 0.0052
     & 0.6825 ± 0.0074
     & 0.6746 ± 0.0056
     \\
     \bottomrule
    \end{tabular}
\end{table}

We run ablation experiments to study the combined effects of patch size vs. model with and without node and patch PE, see Table~\ref{tab: pe and patch size (zinc)} and Table~\ref{tab: pe and patch size (peptide)}.

Overall, increasing the number of patches improves the results independently of using or not the PEs for ZINC and Peptide-func. Node PE clearly helps more than patch PE for both datasets and using both PEs is generally more helpful for a larger number of patches.

\section{Study on Different Designs of Graph-Based MHA}\label{section: design of MHA}

\begin{table}[!ht]
    \centering
    \caption{Different designs of graph-based multi-head attention (gMHA) in transformer layer.}
    \label{tab: graph vit}
    \small
    \begin{tabular}{llcc}
    \toprule
         gMHA &  Equation & ZINC (MAE$\downarrow$)& Peptides-func (AP$\uparrow$) \\
         \midrule
         Standard/Full attention~\citep{vaswani2017attention}
         & $\textrm{softmax}\big(\frac{QK^T}{\sqrt{d}}\big)V$
         & 0.1784 ± 0.0238
         & 0.6778 ± 0.0039 \\
        Graph Attention~\citep{dwivedi2021generalization}
        & $\textrm{softmax}\big(A^P\odot \frac{QK^T}{\sqrt{d}}\big)V$
        & 0.1527 ± 0.0067
        & 0.6795 ± 0.0070  \\
        Kernel Attention~\citep{mialon2021graphit}
        & $\textrm{softmax}\big(\textrm{RW(}A^P)\odot \frac{QK^T}{\sqrt{d}}\big)V$
        & 0.1010 ± 0.0031
        & 0.6844 ± 0.0102 \\
        Additive Attention~\citep{ying2021graphormer}
        & $\textrm{softmax}\big(\frac{QK^T}{\sqrt{d}}\big)V+\textrm{LL}(A^P)$
        & 0.1632 ± 0.0063
        & 0.6842 ± 0.0057\\
        Hadamard Attention
        & $\big(A^P\odot\textrm{softmax}(\frac{QK^T}{\sqrt{d}})\big)V$ 
        & 0.0849 ± 0.0047
        & 0.6919 ± 0.0085 \\
        \bottomrule
    \end{tabular}
\end{table}

We conducted experiments on ZINC and Peptides-func datasets to explore five different versions of Graph ViT. The versions primarily differ in the attention function used. The attention functions we considered are as follows:
(1) Standard/Full Attention: This attention function is based on the original attention mechanism introduced by~\citet{vaswani2017attention}.
(2) Graph Attention: This attention function is derived from the Graph Transformer (GT) model proposed by~\citet{dwivedi2021generalization}.
(3) Kernel Attention: This attention function is based on the kernel attention mechanism proposed by~\citet{mialon2021graphit} in the GraphiT model.
(4) Additive Attention: This attention function is derived from the Graphormer model proposed by~\citet{ying2021graphormer}.
(5) Hadamard Attention: We employed Hadamard attention as the default attention function in our Graph ViT model. 
Results are presented in the Table~\ref{tab: graph vit}.

Experiments clearly demonstrate that the choice of the self-attention function is important. The Hadamard attention provides the best performance for ZINC (0.0849) and for peptides-func (0.6919) among all attention functions.

\section{Effect of Data Augmentation}\label{app sec: da}
\begin{table}[!ht]
\caption{Effect of data augmentation (DA): \xmark \ means no DA and \cmark uses DA.}
\footnotesize
    \centering
    \begin{tabular}{lcccccc}
    \toprule
         \multirow{2}{*}{Model} & \multirow{2}{*}{DA}
         &  \multicolumn{2}{c}{ZINC} &  \multicolumn{2}{c}{Peptides-struct} \\
         \cmidrule(lr){3-4}\cmidrule(lr){5-6}
         & & MAE $\downarrow$ & Time (S/Epoch) & MAE $\downarrow$ & Time (S/Epoch) \\
         \midrule
         \multirow{2}{*}{GCN-MLP-Mixer} 
         & \xmark 
         & 0.2537 ± 0.0139 & 5.3603
         & 0.2761 ± 0.0041 & 6.8297\\
          & \cmark 
          & 0.1347 ± 0.0020 & 5.6728
          & 0.2486 ± 0.0041 & 9.2561\\
          \midrule
          
          \multirow{2}{*}{GatedGCN-MLP-Mixer} 
         & \xmark 
         & 0.2121 ± 0.0172 & 5.3816
         & 0.2776 ± 0.0020 & 7.8609\\
          & \cmark 
          & 0.1244 ± 0.0053 & 5.7786
          & 0.2508 ± 0.0007 & 9.5830\\
          \midrule
          
          \multirow{2}{*}{GINE-MLP-Mixer} 
         & \xmark 
         & 0.1389 ± 0.0171 & 5.3905
         & 0.2792 ± 0.0043 & 7.8849\\
          & \cmark 
          & 0.0733 ± 0.0014 & 5.6704
          & 0.2494 ± 0.0007 & 8.8136\\
          \midrule
          
          \multirow{2}{*}{GraphTrans-MLP-Mixer} 
         & \xmark 
         & 0.1665 ± 0.0145 & 6.0039
         & 0.2802 ± 0.0030 & 9.0999\\
          & \cmark 
          & 0.0773 ± 0.0030 & 6.1616
          & 0.2480 ± 0.0013 & 9.7730\\
         \bottomrule
    \end{tabular}
    \label{app tab: Data Augmentation}
\end{table}

Then proposed data augmentation (DA) corresponds to newly generated graph patches with METIS at each epoch, while no DA means patches are only generated at the initial epoch and then reused during training. Table~\ref{app tab: Data Augmentation} presents different results. First, it is clear that DA brings an increase in performance. Second, re-generating graph patches only add to a small amount of training time. 

It is worth noting that the drop edge technique we use here is different to the standard data augmentation techniques such as DropEdge~\citep{rong2019dropedge}, and G-Mixup~\citep{han2022gmixup}, which either add slightly modified copies of existing data or generate synthetic based on existing data. Our approach is different and actually specific to the Graph MLP-Mixer model.

\section{Long Range Graph Benchmark}\label{app sec: lrgb}
\begin{table}[!ht]
    \caption{Comparison of our best results from Table~\ref{tab: performance} with the state-of-the-art Models on large real world datasets~\citep{dwivedi2022long}.}
\scriptsize
    \centering
    \begin{tabular}{lccccccc}
    \toprule
         \multirow{2}{*}{Model} & \multirow{2}{*}{\# Params}
         &  \multicolumn{3}{c}{Peptide-func}
         &  \multicolumn{3}{c}{Peptide-struct} \\
         \cmidrule(lr){3-5} \cmidrule(lr){6-8}
         & & Avg. Precision  $\uparrow$ & Time (S/Epoch) & Memory (MB)
         & MAE $\downarrow$ & Time (S/Epoch) & Memory (MB)\\
         \midrule
         GCN & 508k
         & 0.5930 ± 0.0023 & 4.59 & 696
         & 0.3496 ± 0.0013 & 4.51 & 686\\
         GINE & 476k 
         & 0.5498 ± 0.0079 & 3.94 & 659
         & 0.3547 ± 0.0045 & 3.84 & 658 \\
         GatedGCN & 509k 
         & 0.5864 ± 0.0077 & 5.48 & 1,038
         & 0.3420 ± 0.0013 & 5.31 & 1,029\\
         GatedGCN + RWSE  & 506k 
         & 0.6069 ± 0.0035 & 5.75 & 1,035
         & 0.3357±  0.0006 & 5.61 & 1,038\\
         \midrule
         Transformer + LapPE & 488k 
         & 0.6326 ± 0.0126 & 9.74 ($1.1\times$) & 6,661 ($6.6\times$)
         & 0.2529 ± 0.0016 & 9.61 ($1.1\times$) & 6,646 ($8.0\times$)\\
         SAN + LapPE~\citep{chen2022structure_SAT} & 493k 
         & 0.6384 ± 0.0121 & 80.47 ($9.4\times$) & 12,493 ($12.4\times$) 
         & 0.2683 ± 0.0043 & 79.41 ($8.8\times$) &  12,226 ($14.7\times$)\\
         SAN + RWSE~\citep{chen2022structure_SAT} & 500k 
         & 0.6439 ± 0.0075 & 68.44 ($8.0\times$) & 19,691 ($19.5\times$)
         & 0.2545 ± 0.0012 & 70.39 ($7.8\times$) &12,111 ($14.5\times$)\\
         GPS~\citep{rampavsek2022recipe} & 504k
         & 0.6562 ± 0.0115 & 11.83 ($1.4\times$) & 6,904 ($6.8\times$)
         & 0.2515 ± 0.0012 & 11.74 ($1.3\times$)& 6,878 ($8.3\times$)\\
         
         \midrule
         GNN-AK+~\citep{zhao2021stars} & 631k
         &  0.6480 ± 0.0089 & 22.52 ($2.6\times$) & 7,855 ($7.8\times$)
         & 0.2736 ± 0.0007  & 22.11 ($2.5\times$) & 7,634 ($9.2\times$)\\
         SUN~\citep{sun} & 508k
         & 0.6730 ± 0.0078 & 376.66 ($43.8\times$) & 18,941 ($18.8\times$)
         & 0.2498 ± 0.0008 &  384.26 ($42.7\times$)& 17,215 ($20.7\times$)\\
         
         \midrule
         GCN-MLP-Mixer & {329k}
         & {0.6832 ± 0.0061} & {8.48} & {716}
         & {0.2486 ± 0.0041} & {8.12} & {679} \\
         
         GatedGCN-MLP-Mixer & {527k} 
         & {0.6932 ± 0.0017} & {8.96} & {969}
         & {0.2508 ± 0.0007} & {8.44} & {887} \\
         
         GINE-MLP-Mixer & {397k}
         & \bf{0.6970 ± 0.0080} & {8.59} ($1.0\times$) & {1,010} ($1.0\times$)
         & {0.2494 ± 0.0007} & {8.51} & {974} \\
         
         GraphTrans-MLP-Mixer & {593k}
         & {0.6858 ± 0.0062} & {9.94} & {975}
         & {0.2480 ± 0.0013} & {9.00} & {1,048}\\

         \midrule
         {GCN-ViT} & {493k}
         & {0.6855 ± 0.0049} & {8.90} & {628 }
         & {0.2468 ± 0.0015} & {8.55} & {609} \\
         
         {GatedGCN-ViT} & {692k} 
         & {0.6942 ± 0.0075} & {9.07} & {848}
         & \bf{0.2465 ± 0.0015} & {9.00} ($1.0\times$) & {833} ($1.0\times$)\\
         
         {GINE-ViT} & {561k}
         & {0.6919 ± 0.0085} & {8.98} & {920} 
         & {0.2449 ± 0.0016} & {8.77} & {902}\\
         
         {GraphTrans-ViT} & {757k}
         & {0.6876 ± 0.0059} & {9.94}  & {975}
         & {0.2455 ± 0.0027} & {9.58} & {981}\\
         \bottomrule 
    \end{tabular}
    \label{app tab: lrgb}
\end{table}

We have provided additional experiments with the recent Long Range Graph Benchmark (LRGB)~\citep{dwivedi2022long} to demonstrate that Graph MLP-Mixer is able to capture long-range interactions. In LRGB, Peptides-func and Peptides-struct are two graph-level prediction datasets, consisting of 15,535 graphs with a total of 2.3 million nodes. The graphs are one order of magnitude larger than ZINC, MolTOX21 and MolHIV with 151 nodes per graph on average and a mean graph diameter of 57. As such, they are better suited to evaluate models enabled with long-range dependencies, as they contain larger graphs and more data points. The performance is reported in Table~\ref{tab: performance}, Table~\ref{tab: sota} and in Table~\ref{app tab: lrgb}. 

We summarize the main results as follows: 1) Graph MLP-Mixer sets new SOTA performance with the best scores of 0.6970 on Peptides-fun and 0.2449 on Peptides-struct (Table~\ref{tab: sota}), demonstrating the ability of the model to better capture long-range relationships.
2) Compared with MP-GNNs (Table~\ref{tab: performance}), Graph MLP-Mixer significantly outperforms the base MP-GNNs; we can observe an average 0.056 Average Precision improvement on Peptides-func and an average 0.028 MAE decrease on Peptides-struct, which verifies its superiority over MP-GNNs in capturing long-range interaction.
3) Graph MLP-Mixer provides significantly better speed/memory complexity compared to Graph Transformer and expressive gnn models, epspecially when training with large graphs, such as SAN+LSPE~\citep{chen2022structure_SAT} and SUN~\citep{sun}. For example, SUN gives similar performance to Graph MLP-Mixer, 0.6730 on Peptides-func and 0.2498 on Peptides-struct, but requires 44x memory and 19x training time (Table~\ref{app tab: lrgb}).

\section{Mitigating Oversquashing in TreeNeighbourMatch}\label{app_sec: oversquashing}

As discussed in section \ref{sec: oversquashing}, 
each example of TreeNeighbourMatch is a binary tree of depth $r$. The goal is to predict an alphabetical label for the target node, where the correct answer is the label of the leaf node that has the same degree as the target node. Figure~\ref{fig: tree neighbour} shows that standard MP-GNNs (i.e., GCN, GGCN, GAT and GIN) fail to generalize on the dataset from $r=4$, whereas
our model mitigates over-squashing and generalizes well until $r=7$.

To better understand these empirical observations, we first note that as shown by \citet{alon2020bottleneck}, MP-GNNs are fundamentally limited in their ability to solve larger TreeNeighbourMatch cases as they `squash' information about the graph into the target node's embedding, which can hold a limited amount of information in their floating point representation. Next, we consider Graph MLP-Mixer from an expressiveness point of view, and provide a simple construction to illustrate that it avoids this problem by transmitting long-range information directly without oversquashing. Concretely, consider each node as one patch. Then, Graph MLP-Mixer's Patch Encoder extracts each node's degree and alphabetical label, storing them into the resulting Patch Embeddings. The next Token Mixer layer then compares each node's degree to the target node's, and outputs an indicator variable for whether these degrees are equal, which is transmitted to the next layer. Finally, by combining each node's alphabetical label and this indicator variable, the Fully Connected layer can then output the alphabetical label of the node with matching degree to the target node. In summary, we can observe that Graph MLP-Mixer can solve TreeNeighbourMatch instances while only requiring that each node embedding to capture information about that patch, not the entire graph, thus avoiding the inherent limitations of MP-GNNs as discussed in \citep{alon2020bottleneck}.

\section{Complexity Analysis}\label{sec: complexity}
For each graph $G = (\mathcal{V}, \mathcal{E})$, with $N=|\mathcal{V}|$ being the number of nodes and $E = |\mathcal{E}|$ being the number of edges, the METIS patch extraction takes $O(E)$ runtime complexity, and outputs graph patches $\{G_1,...,G_P\}$, with $P$ being the pre-defined number of patches. Accordingly, we denote each graph patch as $G_p = (\mathcal{V}_p, \mathcal{E}_p)$, with $N_p=|\mathcal{V}_p|$ being the number of nodes and $E_p = |\mathcal{E}_p|$ being the number of edges in $G_p$.
After our one-hop overlapping adjustment, the total number of nodes and edges of all the patches are $N'=\sum_p N_p \le PN$ and $E' = \sum_p E_p \le PE$, respectively. Assuming base GNN has $O(N+E)$ runtime and memory complexity, our patch embedding module has $O(N'+E')$ runtime and memory complexity, introducing a constant overhead over the base GNN model. For the mixer layers, the complexity is $O(P)$ as discussed in Section~\ref{subsec: mixer layer}.

\section{Limitations}\label{app sec: limitation}
The current limitations of the model are as follows.

\textbf{1) Arbitrary choice of the number of clusters in Metis.} The number of patches needs to be selected and the number is different across different datasets. Besides, selecting the number of patches to be the same for graphs of variable sizes makes the network operate at different levels of graph resolution and may affect the overall performance.

\textbf{2) Empirical experiments on WL-expressivity.} Our results on the expressivity of the Graph MLP-Mixer are empirical. A theoretical analysis of the expressivity of the model on graphs with higher WL degrees would be valuable but such an analysis is non-trivial.

\textbf{3) Training and pre-training on large-scale datasets of small and large graphs.} More experimental results on MalNet~\citep{freitas2020large_malnet}, PascalVOC-SP, COCO-SP~\citep{dwivedi2022long} and PCQM4Mv2~\citep{hu2020open} to further test the supervised ability of the model. Besides, the pre-training capability of Graph MLP-Mixer on large-scale datasets with small graphs and large graphs was also not studied. We leave these tasks as future work.

\end{document}